%% file: main.tex
\documentclass[runningheads]{llncs}

\usepackage{eccv}

\usepackage{eccvabbrv}
\usepackage{booktabs}
\usepackage{multirow} 
\usepackage{verbatim}
\usepackage{cite}
\usepackage{afterpage}
\usepackage{pifont}
\usepackage{graphicx}
\usepackage{booktabs}
\usepackage{amsmath,amsfonts}
\usepackage{algorithmic}
\usepackage[ruled,linesnumbered]{algorithm2e}
\usepackage[colorlinks,
linkcolor=blue,
anchorcolor=blue,
citecolor=blue]{hyperref}
\newcommand{\cmark}{\ding{51}}%
\usepackage{threeparttable}

\usepackage[accsupp]{axessibility}  
\usepackage{xcolor}
\definecolor{strategytext}{RGB}{169, 233, 228}
\definecolor{dgfmtext}{RGB}{227, 212, 231}

\usepackage{hyperref}

\usepackage{orcidlink}

\begin{document}

\title{DyGeoVLN: Infusing Dynamic Geometry Foundation Model into Vision-Language Navigation} 

\titlerunning{DyGeoVLN}

\author{Xiangchen Liu\inst{1}\textsuperscript{*} \and
Hanghan Zheng\inst{2}\textsuperscript{*} \and
Jeil Jeong\inst{1} \and Minsung Yoon\inst{1} \and Lin Zhao\inst{3} \and Zhide Zhong\inst{2} \and Haoang Li\inst{2}\textsuperscript{\dag} \and Sung-Eui Yoon\inst{1}\textsuperscript{\dag}}

\authorrunning{X. Liu et al.}

\institute{$^1$KAIST \quad $^2$HKUST(GZ) \quad $^3$JD Explore Academy}

\renewcommand{\thefootnote}{}
\footnotetext{\textsuperscript{*}Equal Contribution, \textsuperscript{\dag}Corresponding Authors}

\maketitle

\begin{abstract}
Vision-language Navigation (VLN) requires an agent to understand visual observations and language instructions to navigate in unseen environments. Most existing approaches rely on static scene assumptions and struggle to generalize in dynamic, real-world scenarios. To address this challenge, we propose DyGeoVLN, a dynamic geometry-aware VLN framework. Our method infuses a dynamic geometry foundation model into the VLN framework through cross-branch feature fusion to enable explicit 3D spatial representation and visual-semantic reasoning. To efficiently compress historical token information in long-horizon, dynamic navigation, we further introduce a novel pose-free and adaptive-resolution token-pruning strategy. This strategy can remove spatio-temporal redundant tokens to reduce inference cost. Extensive experiments demonstrate that our approach achieves state-of-the-art performance on multiple benchmarks and exhibits strong robustness in real-world environments.
\end{abstract}

\section{Introduction}
\label{sec:intro}
\input{vln_introduction}

\section{Related Works}
\input{vln_relatedwork}

\section{Method}
\input{vln_method}
\section{Experiments}
\input{gdpf_exp}
\section{Conclusion}
In this paper, we presented DyGeoVLN, a dynamic geometry-aware VLN framework that infuses a dynamic geometry foundation model into VLN to achieve explicit dynamic spatial representation alongside strong visual–semantic reasoning. We also proposed an occupancy-aware spatial token pruning strategy to control token growth for efficient long-horizon navigation. Extensive experiments demonstrate that DyGeoVLN achieves state-of-the-art performance on both dynamic and static benchmarks and can be effectively deployed in real-world environments. Ultimately, this work advances the VLN research from static semantic reliance to dynamic, geometry-aware embodied navigation.


%
%

\bibliographystyle{splncs04}
\bibliography{main}

\newpage
\appendix
\noindent\textbf{\Large Appendix}

\vspace{1.0em}

Section~\ref{sec:PRUNE} makes a detailed explanation about the workflow of our adaptive-resolution and occupancy-aware spatial token pruning strategy. Section~\ref{sec:VGFM} presents further insight into our dynamic geometry foundation model and the DyHM3D dataset. Section~\ref{sec:benchmark} illustrates qualitative navigation performance on both dynamic and static VLN benchmarks. Section~\ref{sec:realworld} details our real-world deployment and experimental results.

\section{Work Flow of Our Spatial Token Pruning Strategy} \label{sec:PRUNE}
 The Algorithm~\ref{alg:adaptive-pruning} describes the workflow of our token pruning strategy, which corresponds to our section ``3.3 Adaptive-resolution and occupancy-aware spatial token pruning'' in the manuscript.
\begin{algorithm}[!ht]
\caption{Adaptive-resolution and occupancy-aware token pruning}
\label{alg:adaptive-pruning}
\begin{algorithmic}[1]
\REQUIRE Local points $\{\mathbf{P}_t\}$, poses $\{\mathbf{T}_t\}$, fused tokens $\{\mathbf{F}_t\}$, 
          voxel size $\Delta$, ratio $\rho$, selection rule $\mathsf{Sel}(\cdot)$
\ENSURE Pruned token sets $\{\mathbf{F}_t^{\text{pruned}}\}$
\FOR{each historical frame $t$}
    \STATE Project points to world coordinates: $\mathbf{X}_t \leftarrow \mathbf{T}_t \mathbf{P}_t$
    \STATE Quantize $\mathbf{X}_t$ into adaptive voxel indices $\mathbf{v}_t$ using $\Delta_t$
\ENDFOR
\STATE Group all tokens $\{\mathbf{F}_t\}$ by shared voxel index
\FOR{each voxel $u$ with token set $\mathcal{T}_u$}
    \STATE Select representative tokens $\mathcal{S}_u \leftarrow \mathsf{Sel}(\mathcal{T}_u)$
\ENDFOR
\STATE Mark all tokens in $\bigcup_u \mathcal{S}_u$ as initially preserved
\FOR{each frame $t$}
    \STATE If token ratio $< \rho$, compute importance scores for discarded tokens
    \STATE Add highest-scoring discarded tokens until reaching the ratio $\rho$
    \STATE Smooth the binary mask over a temporal window
    \STATE Collect final preserved tokens $\mathbf{F}_t^{\text{pruned}}$
\ENDFOR
\RETURN $\{\mathbf{F}_t^{\text{pruned}}\}$
\end{algorithmic}
\end{algorithm}

\clearpage
\section{Dynamic Geometry Foundation Model} \label{sec:VGFM}


\subsection{Visualization of the DyHM3D Dataset}

\begin{figure}[!ht]
\centering
\includegraphics[width=0.8\linewidth]{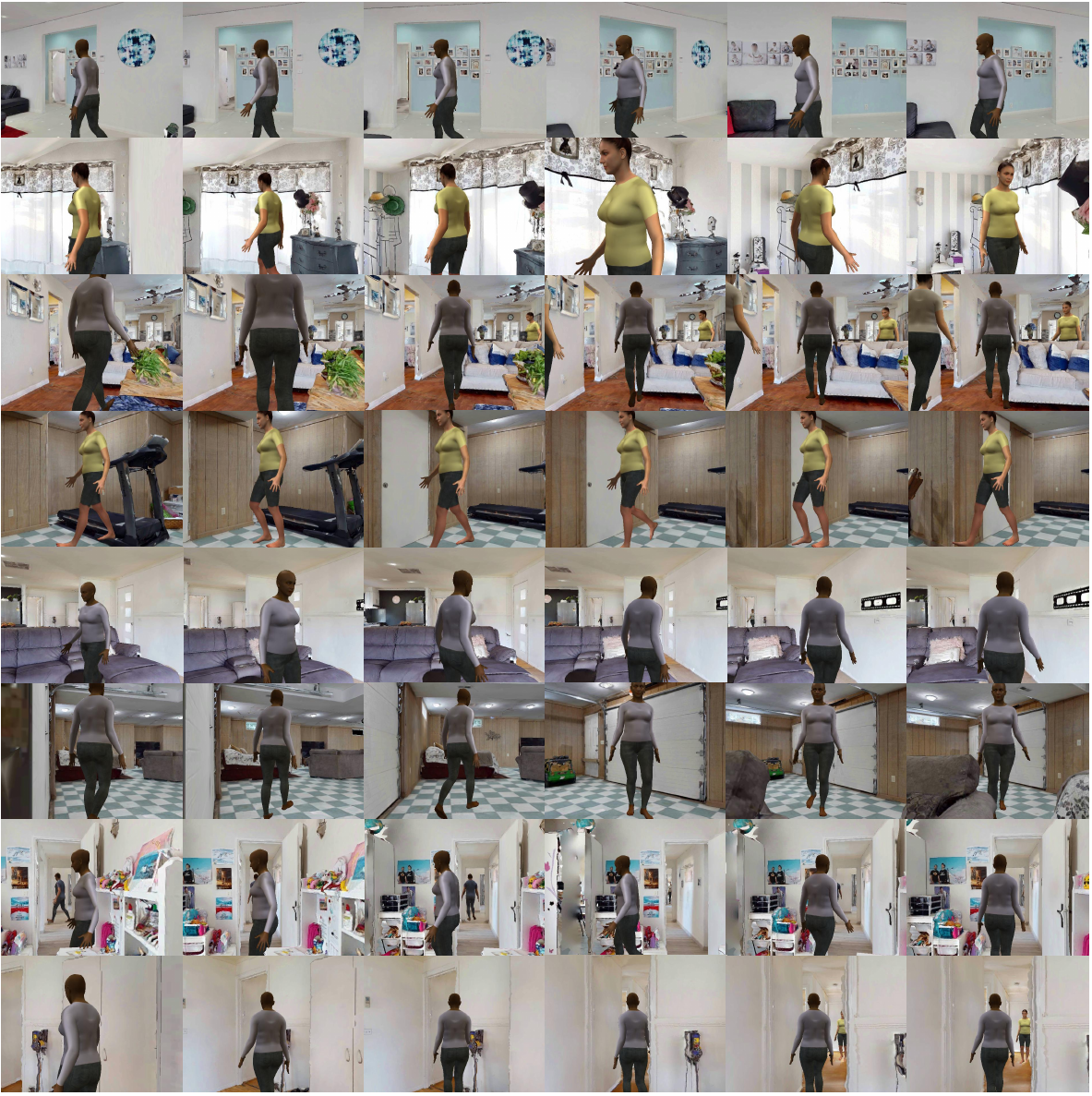}
\caption{Example cases of our DyHM3D dataset.}
\label{fig:DynHM3D_dataset}
\end{figure}

We provide additional qualitative examples of the proposed DyHM3D dataset. As shown in Fig.~\ref{fig:DynHM3D_dataset}, the dataset contains diverse indoor scenes with simulated human motions following navigable trajectories. 

\subsection{Extended Comparison of Reconstruction Performance}
\begin{figure}[!ht]
\centering
\includegraphics[width=\linewidth]{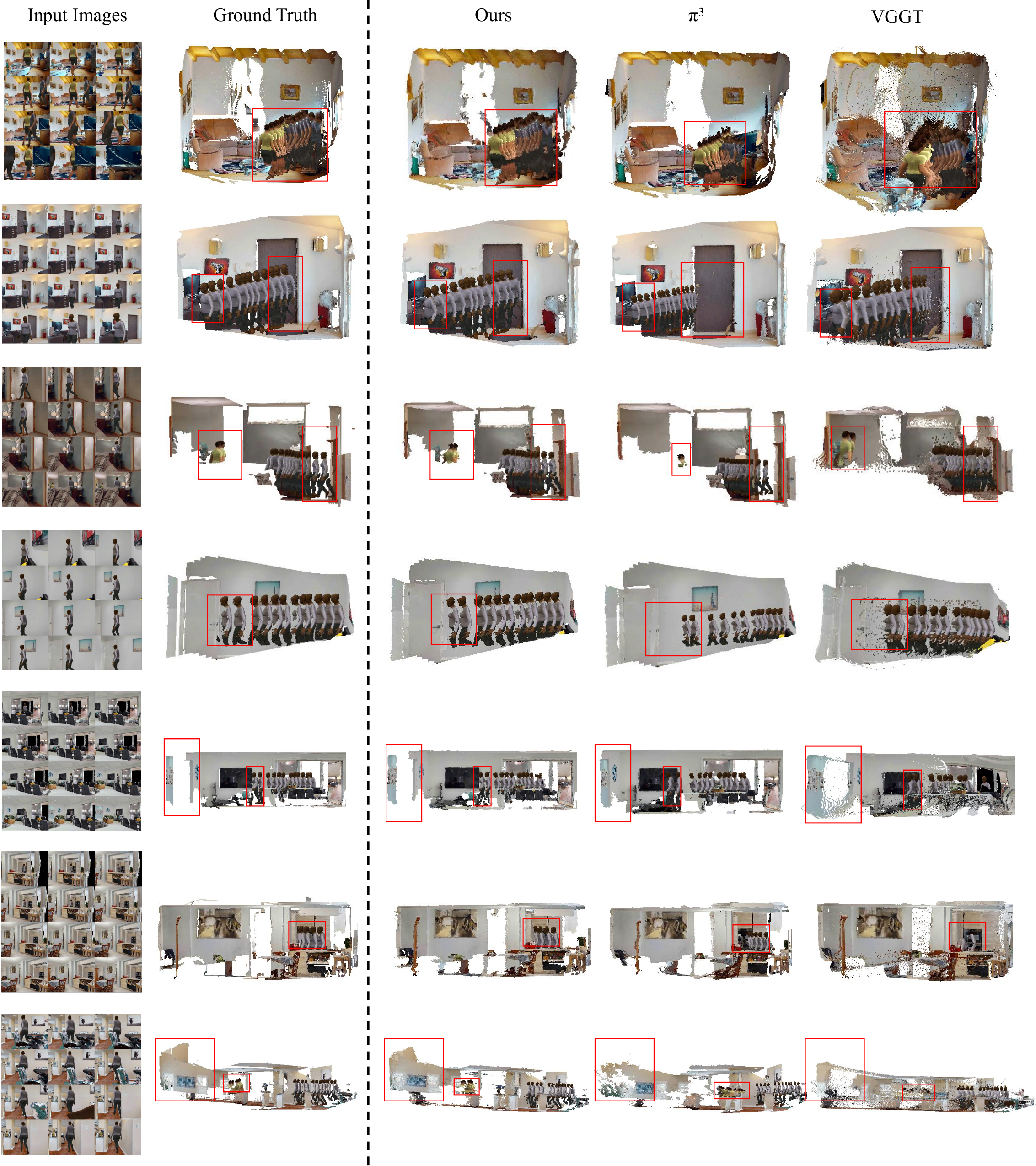}
\caption{Qualitative comparison of reconstruction performance on dynamic scenes.}
\label{fig:dynamic-aware_VGFM}
\end{figure}

To further evaluate the reconstruction quality, Figure~\ref{fig:dynamic-aware_VGFM} illustrates an extended qualitative comparison among $\pi^3$~\cite{wang2025pi3}, VGGT~\cite{wang2025vggt}, and our proposed DGFM. As observed, our model outperforms $\pi^3$ and VGGT in dynamic human perception while maintaining competitive geometric completeness. In dynamic scenes, $\pi^3$ and VGGT often exhibit spatial localization inaccuracies, and the reconstructed human figures frequently suffer from noticeable geometric distortions. In contrast, our DGFM demonstrates strong robustness, consistently producing clear and structurally coherent human geometries that closely align with the ground truth. In addition, our model yields a more uniform point cloud distribution in textureless regions and distant areas.

\section{Qualitative Results on Simulation Benchmarks} \label{sec:benchmark}

\subsection{Visualizations on Dynamic HA-VLN Benchmark}
Figure \ref{fig:havlnr2r} illustrates the navigation results of our proposed DyGeoVLN model within the dynamic HA-VLN~\cite{dong2025ha} benchmark. 
\begin{figure*}[ht]
\centering
    \includegraphics[width=\linewidth]{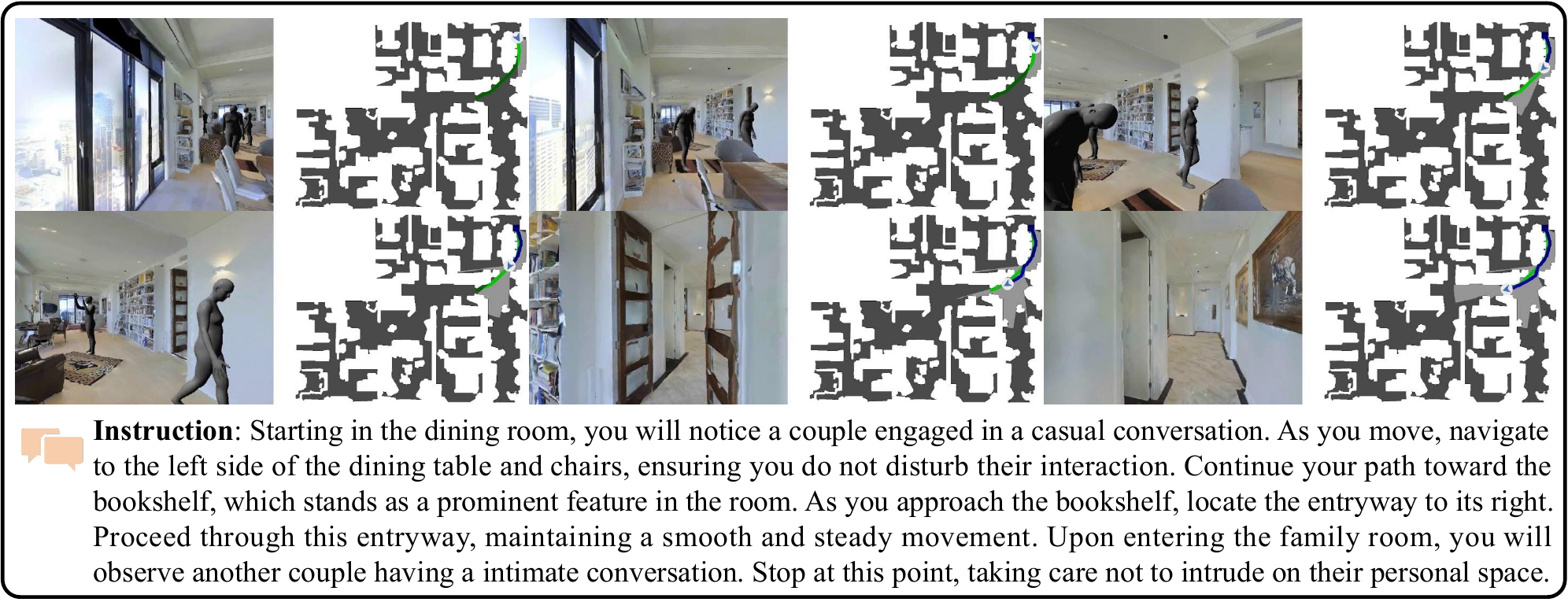}
    \vspace{0.1em} 
    \includegraphics[width=\linewidth]{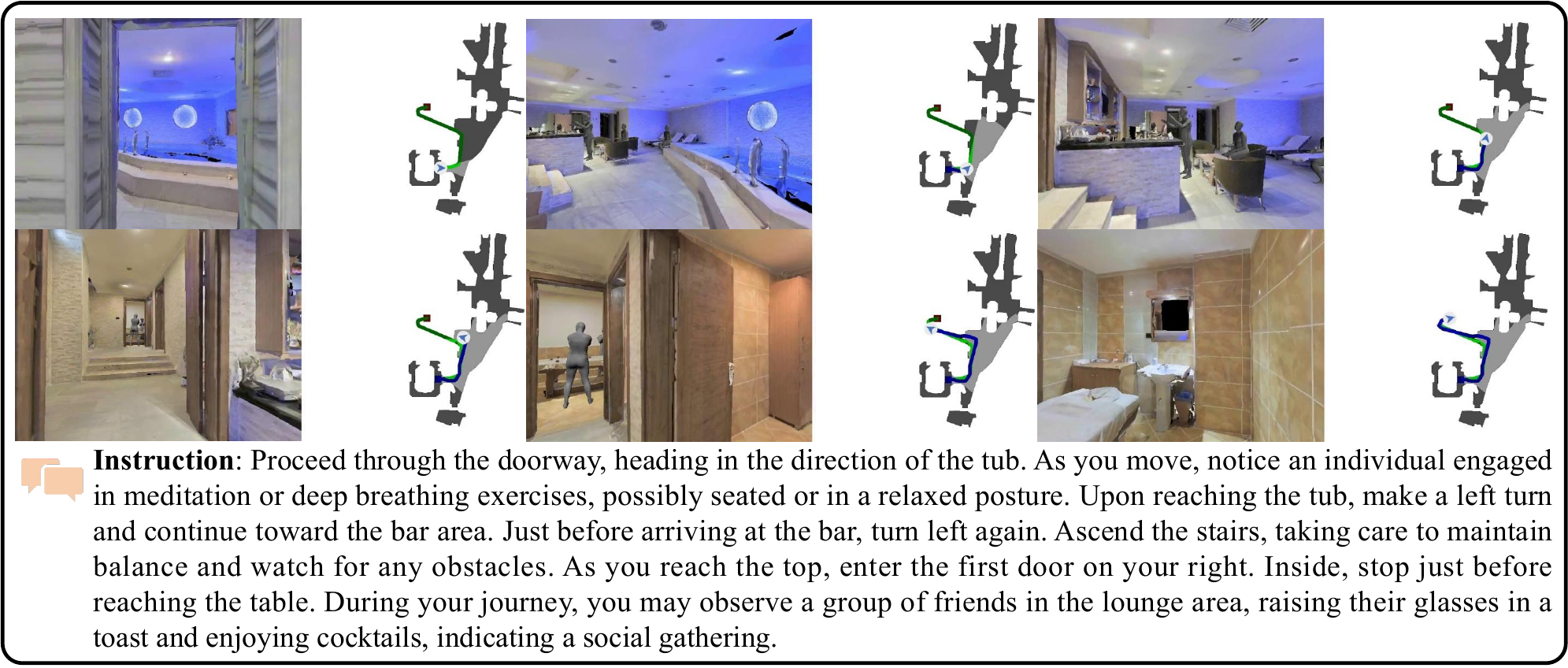}
    \caption{Qualitative results of our DyGeoVLN model on dynamic HA-VLN~\cite{dong2025ha} benchmark. (Cases 1-2)}
    \label{fig:havlnr2r}
\end{figure*}

\begin{figure*}[ht]
    \ContinuedFloat 
    \centering
    \includegraphics[width=\linewidth]{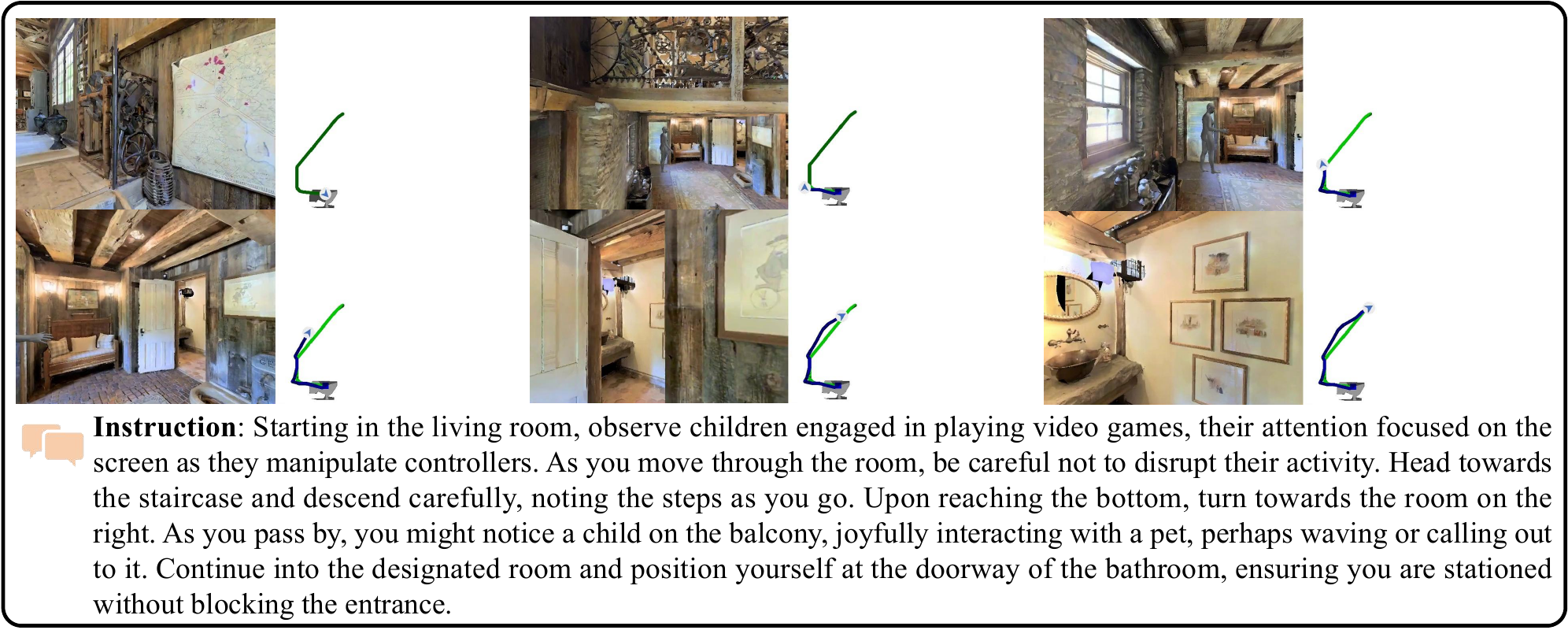}
    \vspace{0.1em}
    \includegraphics[width=\linewidth]{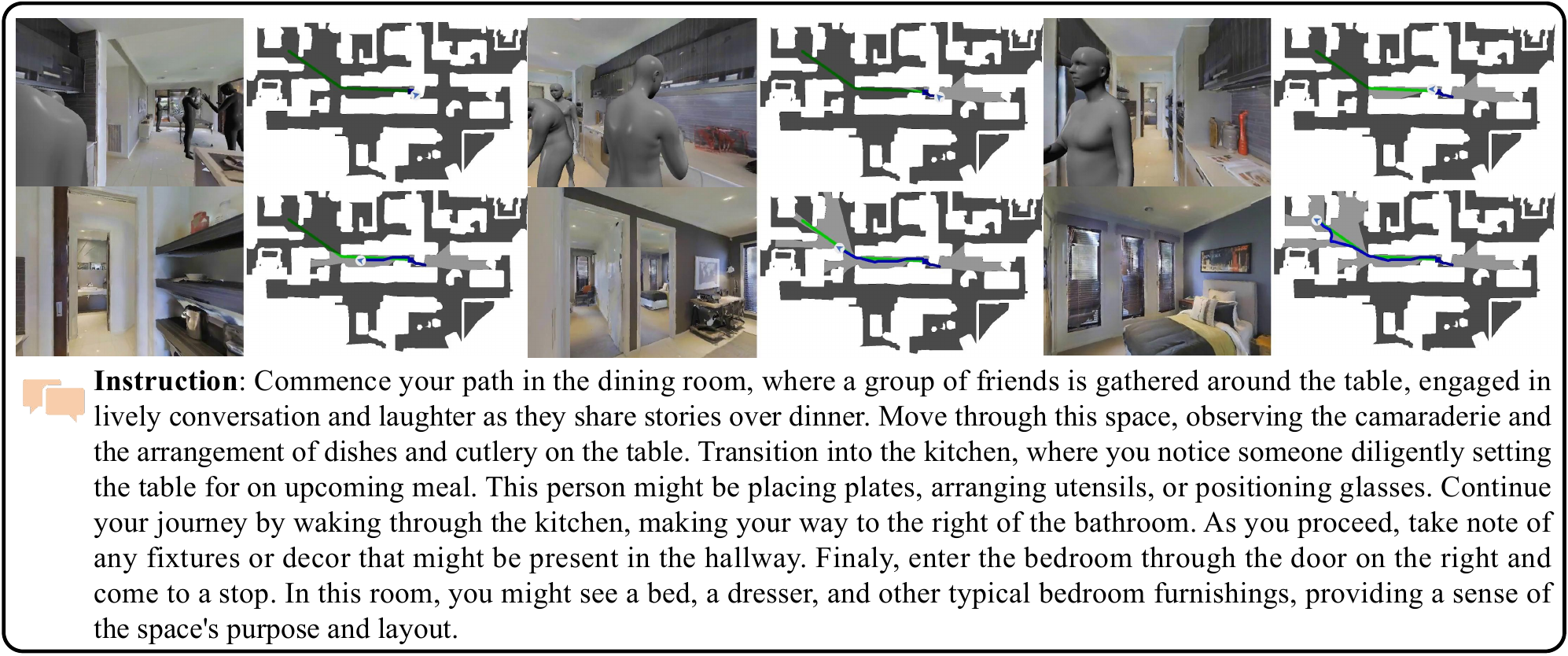}
    \vspace{0.1em}
    \includegraphics[width=\linewidth]{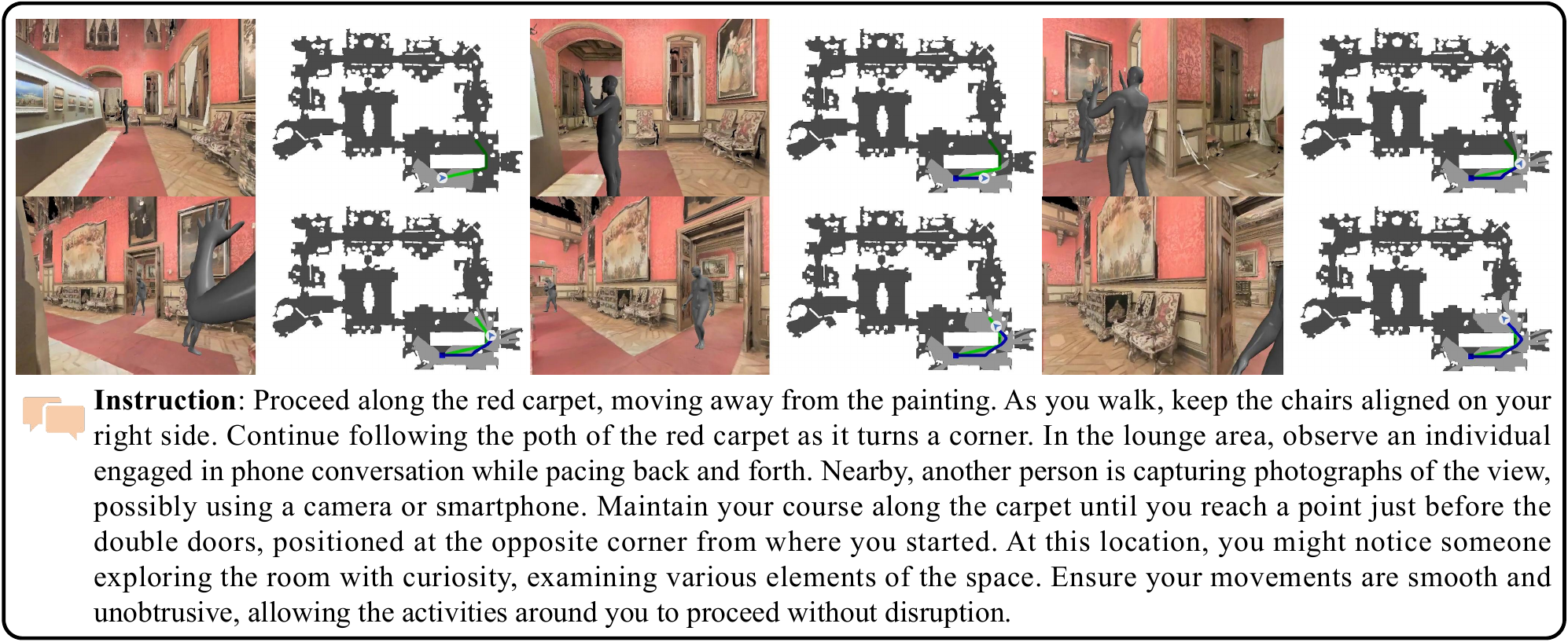}
    \caption[]{Qualitative results of our DyGeoVLN model on dynamic HA-VLN~\cite{dong2025ha} benchmark. (Cases 3-5 continued).}
\end{figure*}


\clearpage
\subsection{Visualizations on Static VLN Benchmark}
\begin{figure*}[!ht]
\centering
    \includegraphics[width=\linewidth]{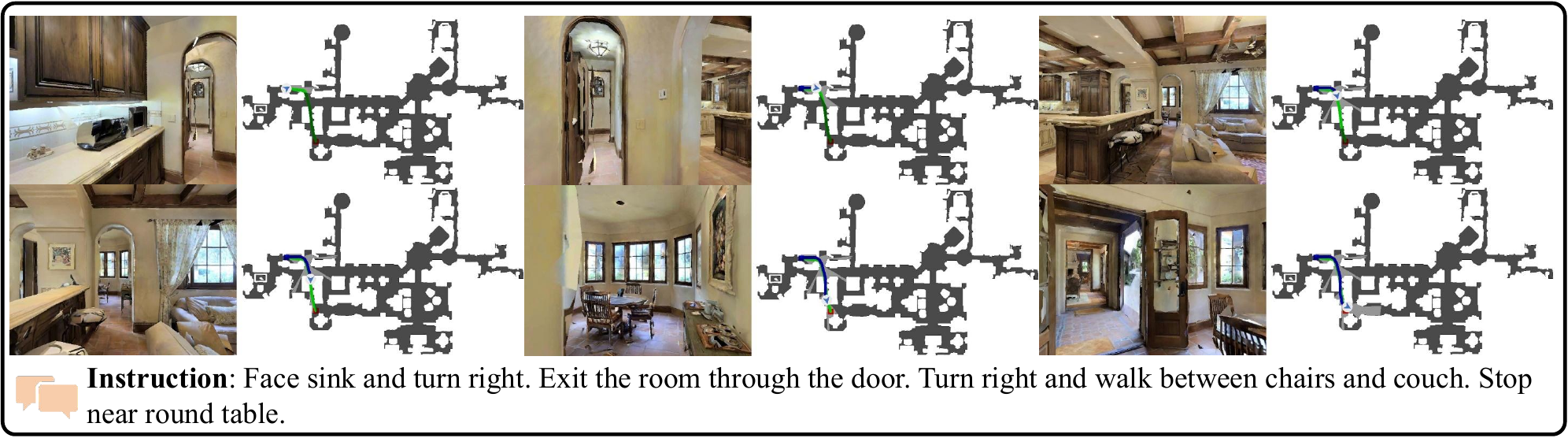}
    \vspace{0.01em} 
    \includegraphics[width=\linewidth]{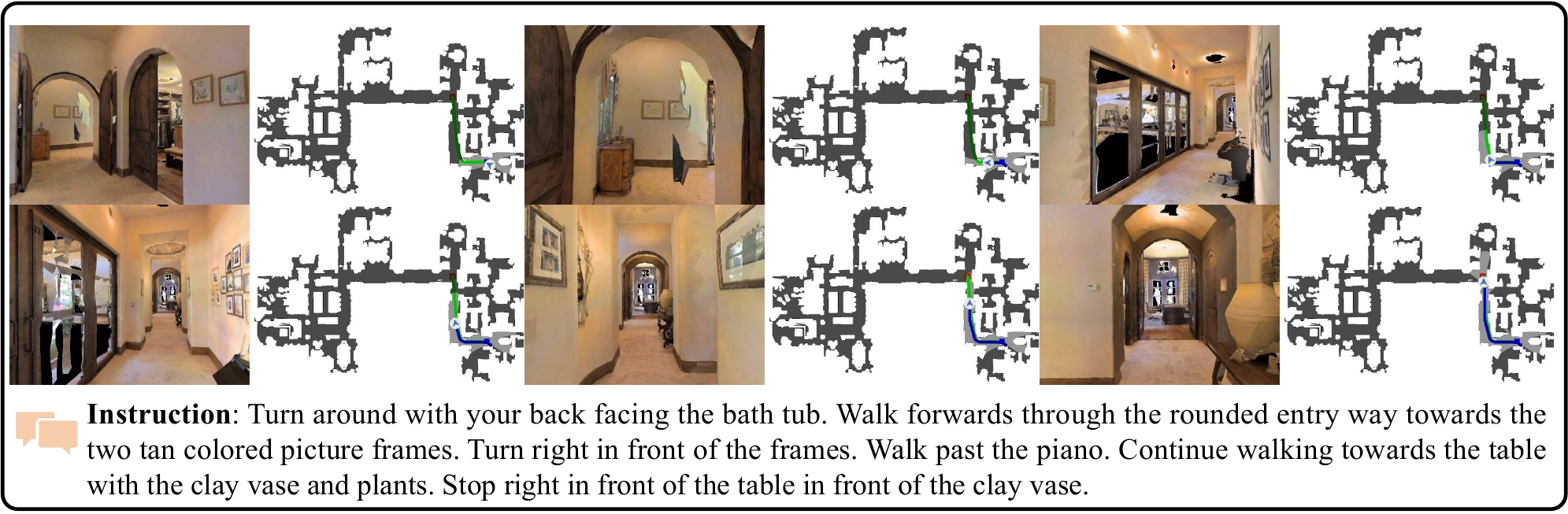}
    \vspace{0.01em}
    \includegraphics[width=\linewidth]{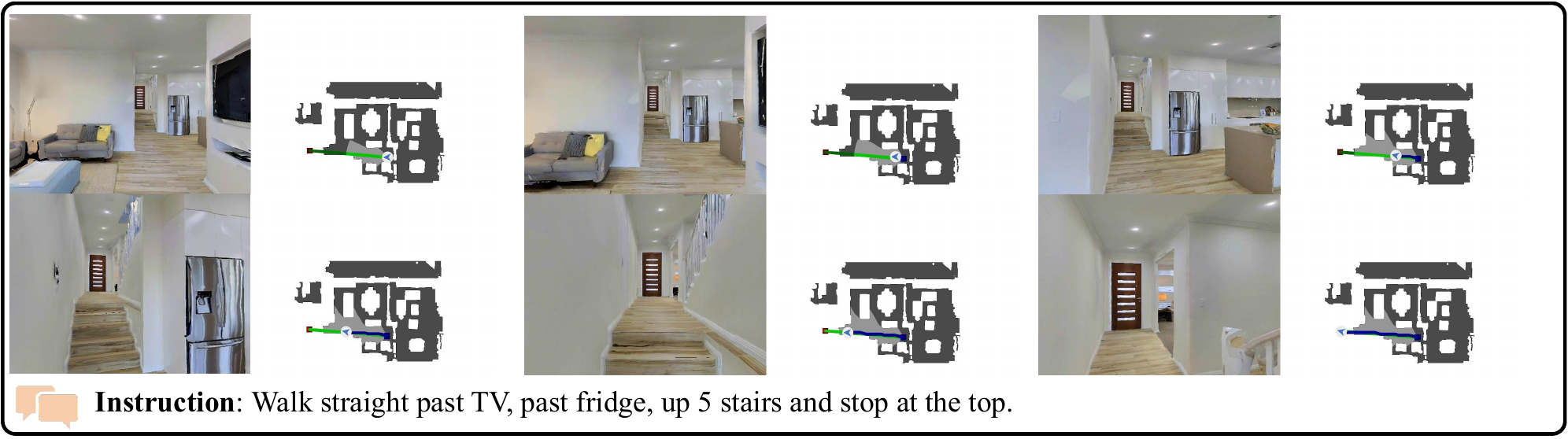}
    \vspace{0.01em}
    \includegraphics[width=\linewidth]{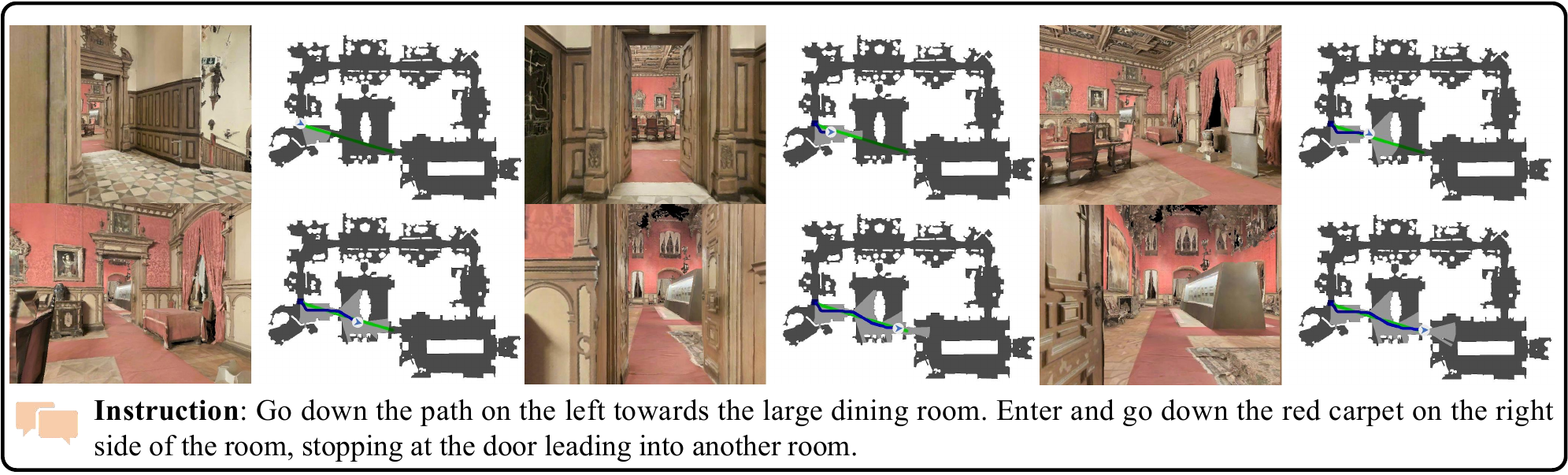}
    \caption{Qualitative results of our DyGeoVLN model on R2R-CE~\cite{krantz2020beyond} benchmark.}
    \label{fig:r2r_qual}
\end{figure*}

Figure \ref{fig:r2r_qual} illustrates the results of our proposed DyGeoVLN model within the static R2R-CE~\cite{krantz2020beyond} VLN benchmark. 

\clearpage
\section{Real-world Experiments} \label{sec:realworld}

\subsection{Real-world Experiment Setup}
\begin{figure}[!h]
\centering
\includegraphics[width=0.9\linewidth]{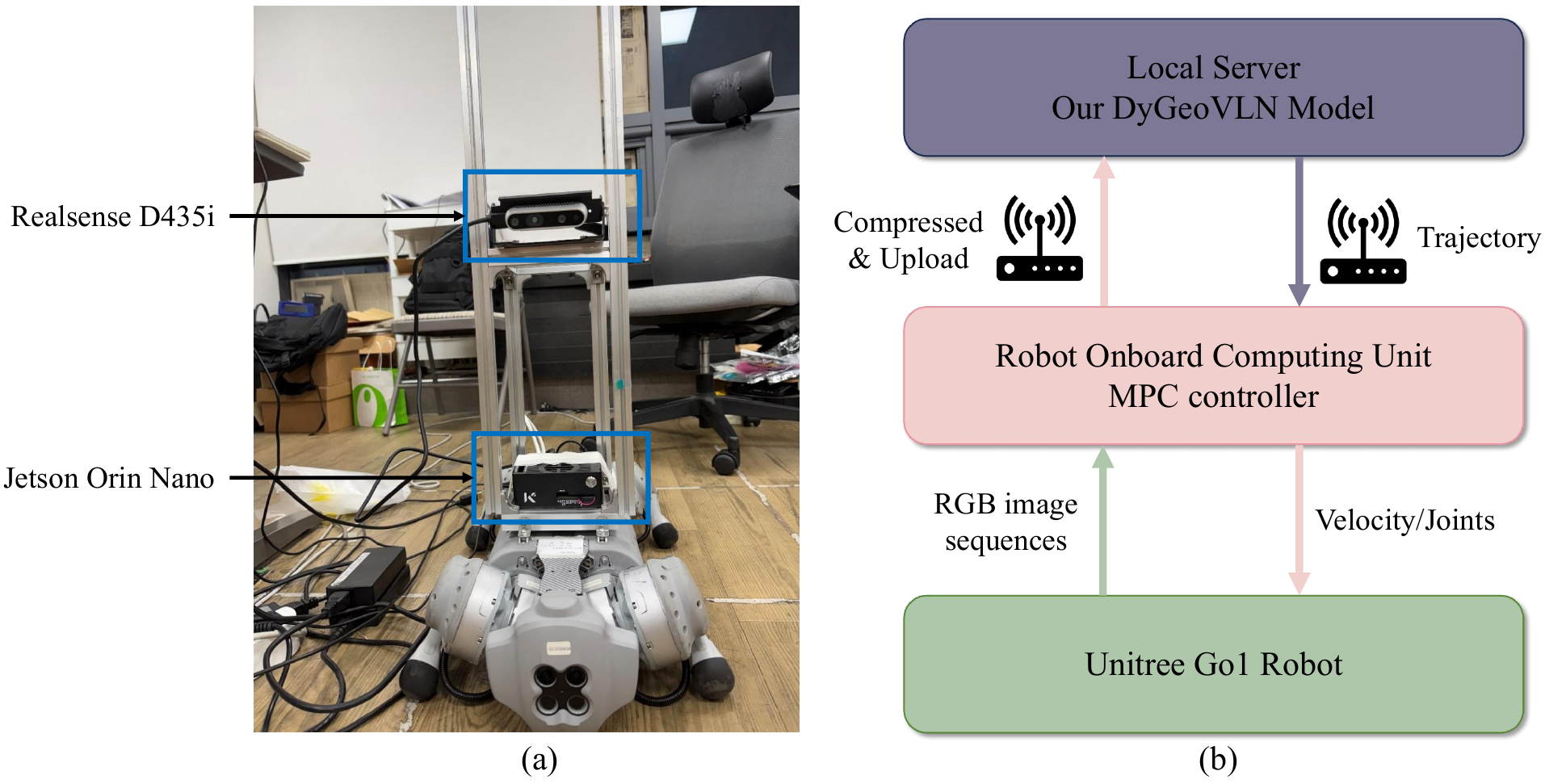}
\caption{Setup of real-world experiments: (a) Hardware configuration of the quadruped robot; (b) Schematic of the real-world deployment pipeline.}
\label{fig:go1steup}
\end{figure}

Our real-world experiments are conducted using a Unitree Go1 quadruped robot. As shown in Figure~\ref{fig:go1steup}, the legged platform is equipped with a forward-facing Intel RealSense D435i camera and an onboard NVIDIA Jetson Orin Nano. The D435i continuously provides video streams, which are received by the Jetson Orin Nano. The Jetson packages the streams into sequences of monocular RGB frames and transmits them via Wi-Fi to a local server, where our DyGeoVLN policy is deployed and runs in real time. For each incoming frame sequence and language instruction, the workstation runs DyGeoVLN and sends back the predicted continuous trajectories to the robot. Onboard the robot, the Orin Nano serves as the interface between planning and control: it receives waypoints, fuses them with the robot's ego state, and generates corresponding low-level control commands, which are transmitted through the SDK interface to drive the robot.

\clearpage
\subsection{Real-world Qualitative Results}
\begin{figure}[!ht]
\centering
\includegraphics[width=0.98\linewidth]{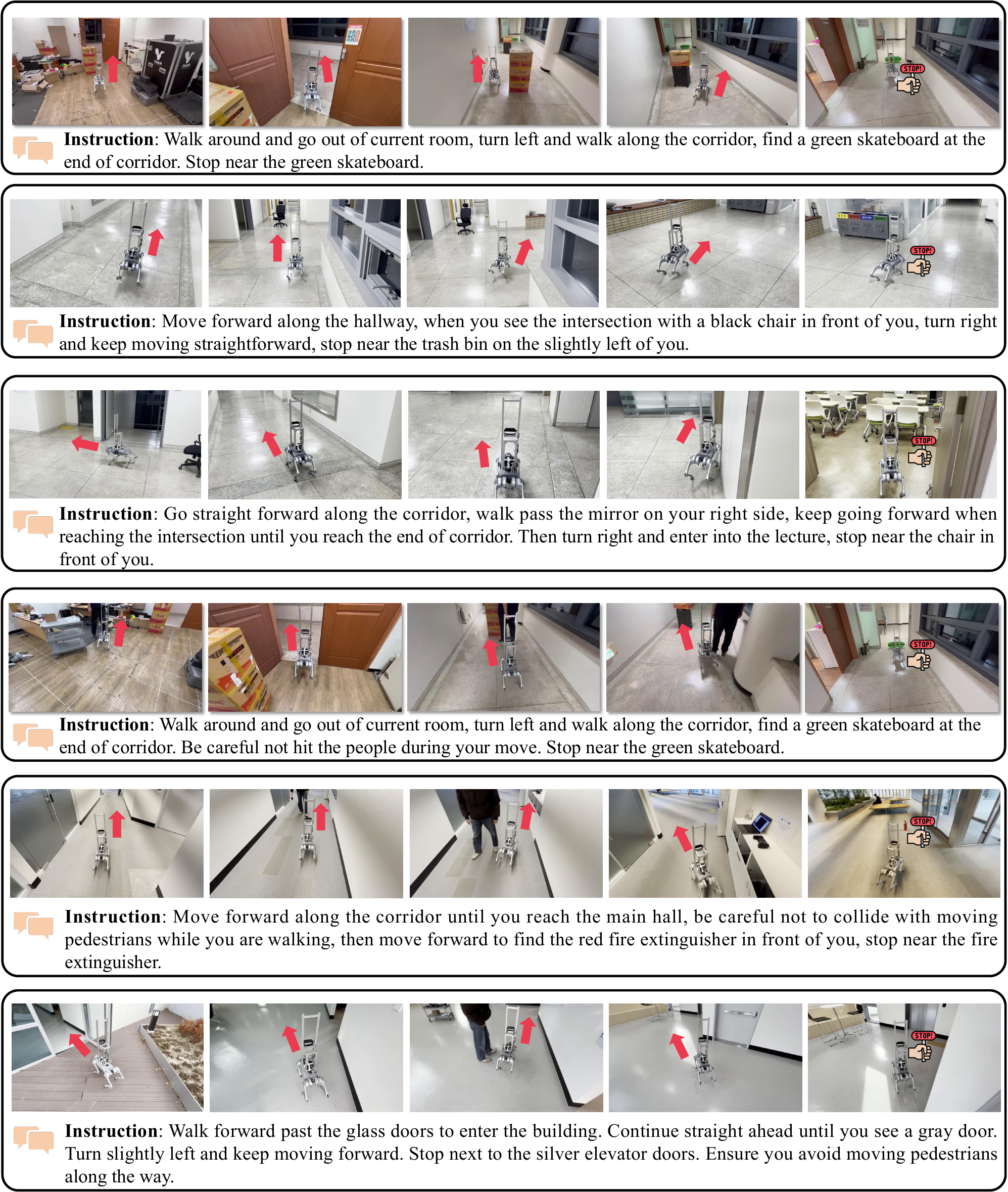}
\caption{Qualitative results of our DyGeoVLN model on real-world experiments: including dynamic and static scenarios.}
\label{fig:realdemo}
\end{figure}

We evaluate our model in several challenging settings, including cluttered rooms and corridors with moving people, as shown in Figure~\ref{fig:realdemo}.

\end{document}

%% file: vln_introduction.tex
In Embodied AI, Vision-Language Navigation (VLN)~\cite{krantz2020beyond,ku2020room} studies agents that rely on egocentric visual observations and language instructions to navigate unseen environments. Recently, advances in Multimodal Large Language Models (MLLMs)~\cite{touvron2023llama,wang2024qwen2,yang2025qwen3} with strong visual perception and semantic reasoning abilities have given rise to a new line of VLN research. Using large-scale pretraining, these models are adapted into vision-language-action architectures that jointly learn visual perception, instruction understanding, and action generation within a unified end-to-end paradigm~\cite{cheng2025navila,zhang2024navid,zhang2024uni,zhang2025embodied}, representing a promising research direction for VLN.

Despite these advances, existing VLN methods face a fundamental problem: navigation is inherently an interaction within a 3D physical world, yet the visual foundations of current systems are primarily built on 2D image-text pre-training~\cite{tschannen2025siglip,radford2021Learning}. Although these 2D foundations provide strong semantic understanding, they severely lack globally consistent 3D spatial reasoning over long-horizon trajectories. To overcome this limitation, recent studies~\cite{zeng2025janusvln,zheng2025efficient} have attempted to incorporate explicit 3D information into MLLMs. However, these approaches typically rely on direct incorporation of features from off-the-shelf 3D foundation models, such as VGGT~\cite{wang2025vggt}, which struggle to effectively handle dynamic environments. Specifically, while these foundation models provide accurate geometric representations for static scenes, their performance degrades substantially in the presence of pronounced scene dynamics.

Beyond the challenge of dynamic 3D spatial reasoning, managing historical token information during navigation is also challenging due to the rapid accumulation of visual data, especially in long-horizon tasks. Some methods alleviate this by down-sampling a fixed number of frames~\cite{cheng2025navila}, thereby reducing the token volume but also discarding fine-grained temporal cues essential for accurate environmental understanding. Other approaches maintain compact memories by pooling or merging visual tokens~\cite{zhang2024uni,zhang2024navid}. Although such strategies are effective at mitigating token explosion, they either fail to fully exploit temporal cues or incur significant computational overhead from continuous reprocessing of historical frames, limiting their scalability for real-world deployment.

\begin{figure}[!t]
\centering
\includegraphics[width=\linewidth]{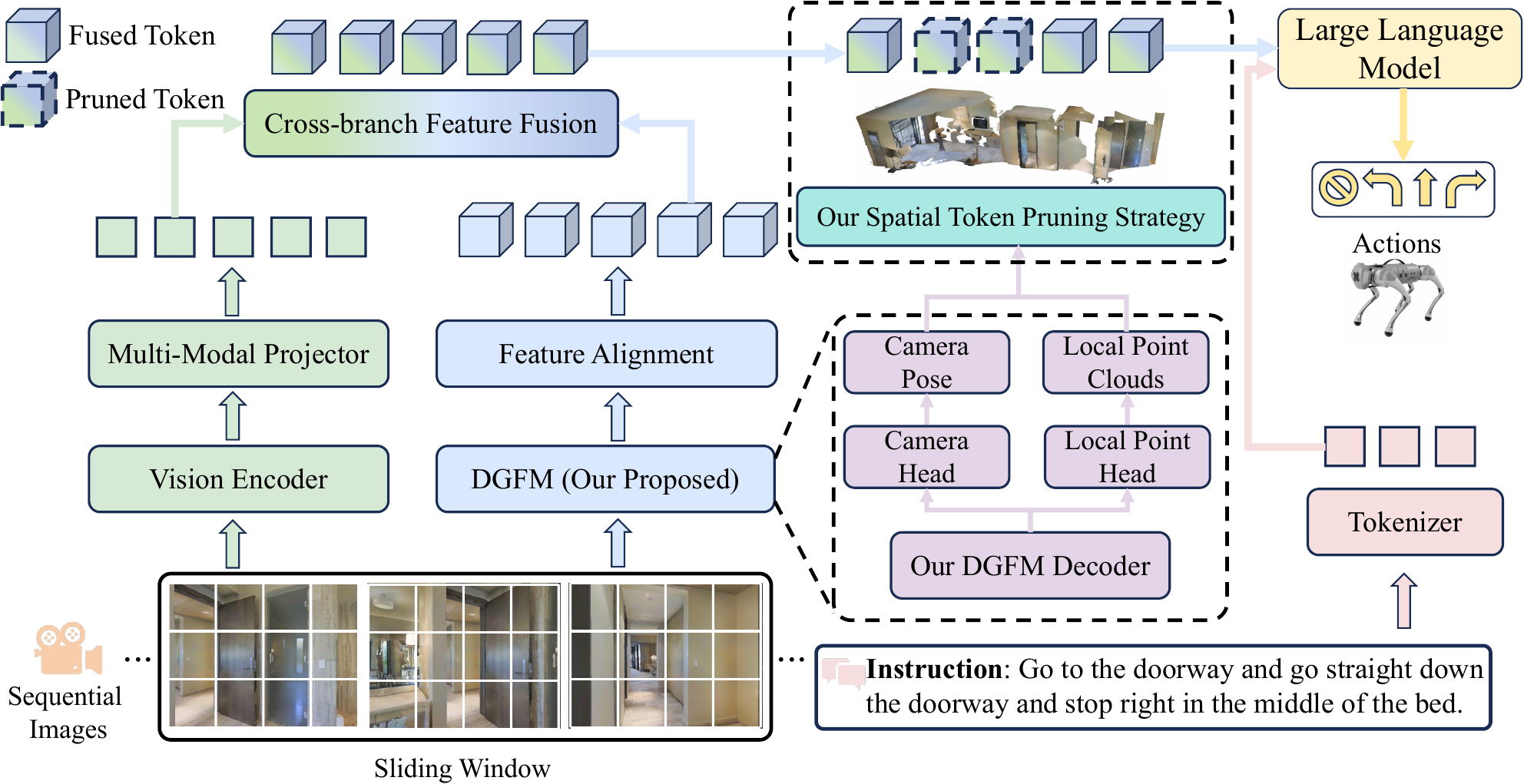}
\caption{The system overview of our DyGeoVLN. Given instruction and sequential images, the vision encoder and our proposed \textbf{\textcolor{dgfmtext}{dynamic geometry foundation model (DGFM)}} respectively produce visual tokens and geometry tokens in a stream manner. Then these tokens are fused via cross-branch feature fusion to obtain spatial-semantic fused tokens. Finally, the LLM integrates the fused tokens and instruction tokens, followed by de-tokenizing the predicted sequence directly into robot actions. To efficiently compress the historical token, we build a \textbf{\textcolor{strategytext}{spatial token pruning strategy}} on top of our \textbf{\textcolor{dgfmtext}{DGFM}}, yielding a compact set of pruned spatial-semantic tokens.}
\label{fig:vlnsys}
\end{figure}

To address the above challenges of dynamic 3D spatial reasoning and historical token information management, this paper presents DyGeoVLN, a framework that infuses a Dynamic Geometry Foundation Model into VLN. As illustrated in Fig.~\ref{fig:vlnsys}, DyGeoVLN constructs a dynamic geometry-aware VLN framework by tightly coupling spatial and semantic information within a unified token space. For solving dynamic spatial reasoning, on the geometry side, we propose a dynamic geometry foundation model that combines a feed-forward geometry foundation model with monocular depth estimation~\cite{yang2024depth}. The explicit injection of these 3D spatial cues into the decoder empowers the model to achieve fine-grained 3D reconstruction and comprehensive dynamic scene representation, ultimately producing robust and aligned geometric representations. To facilitate the training of this geometry model, we further introduce DyHM3D, a dynamic human-centric 3D dataset. On the language side, instruction tokens and fused visual–geometric tokens are jointly processed by the MLLM through cross-branch interaction, enabling explicit 3D global spatial reasoning while preserving rich semantic priors from large-scale pretraining. 

For efficiently compressing historical token information and removing spatial redundant information, we introduce a spatial-aware token-pruning strategy, enabling pose-free and adaptive-resolution redundancy elimination to effectively constrain token growth during long-horizon navigation. Unlike previous voxel-based strategies~\cite{wei2025streamvln} that rely on simulator-provided poses and depth~\cite{savva2019habitat}, our pruning module utilizes the self-inferred poses and point clouds from the geometry model, significantly simplifying the system setup for real-world hardware.

The main contributions of this paper are summarized as follows:
\begin{itemize}
\item We propose DyGeoVLN, a dynamic geometry-aware vision-language navigation framework that infuses the dynamic geometry foundation model into VLN through cross-branch feature fusion, enabling explicit geometry-aware spatial reasoning while preserving strong visual–semantic alignment.
\item We develop a dynamic geometry foundation model that enhances a feed-forward geometry encoder with monocular depth estimation and explicitly injects 3D spatial cues into the decoder, enabling fine-grained 3D reconstruction and dynamic scene representation. For foundation model training, we establish DyHM3D, a novel dynamic human-centric 3D dataset.

\item We further propose a novel occupancy-aware spatial token pruning strategy that leverages dynamic geometry foundation model information, enabling pose-free and adaptive-resolution historical token redundancy reduction.
\end{itemize}

Extensive experiments validate the effectiveness of our approach, with the system achieving state-of-the-art results on multiple benchmarks and demonstrating strong robustness and generalization in dynamic real-world scenarios.

%% file: vln_relatedwork.tex
\subsection{Multimodal Large Language Models for VLN}

The rapid progress of Multimodal Large Language Models (MLLMs)~\cite{touvron2023llama,wang2024qwen2,lin2024video} has sparked a new line of research that treats VLN as a vision-language-action problem. Recent approaches adapt MLLMs to navigation by jointly modeling visual perception, instruction understanding, and action token generation within a unified framework~\cite{cheng2025navila,zhang2024navid,zhang2024uni}. A central challenge in these models is managing the rapidly growing number of visual tokens that arise when processing long egocentric video streams.

Existing methods take several different strategies to manage historical token information. Frame-subsampling approaches~\cite{cheng2025navila} feed only a fixed number of sparsely sampled frames into the MLLM, which reduces memory usage but inevitably discards fine-grained temporal cues that are important for low-level control and dynamic-scene understanding. Memory-based methods~\cite{zhang2024navid,zhang2024uni} pool or merge past tokens into compact summaries, while instruction-centric approaches convert historical observations into textual prompts~\cite{zhou2024navgpt}. Although these designs help limit token growth, they either under-utilize temporal information or introduce substantial redundant computation because past frames must be repeatedly reprocessed. Some work~\cite{wei2025streamvln} reduces token load via voxel-based pruning, but relies on simulator-provided ground-truth poses and depth maps, which are hard to obtain in real-world deployments without additional sensors and localization systems. In contrast, we propose a pose-free, occupancy-aware token pruning strategy that constructs geometry-informed voxel grids from the foundation model itself, enabling direct estimation of pose and point clouds without relying on externally provided pose or depth information, while maintaining a stable token budget during long-horizon navigation. This enables deployment of our model with a minimal sensor setup (a single monocular camera) in both simulation and the real world.

\subsection{Spatial Reasoning with Visual Geometry Foundation Models}

Most current VLN models~\cite{cheng2025navila,wei2025streamvln} use vision encoders~\cite{radford2021Learning,tschannen2025siglip} that are pre-trained on large-scale 2D image–text pairs. Although these encoders provide strong semantic understanding for visual observations, they inherently lack geometric priors and struggle to reason about spatial information across continuous trajectories. As a result, maintaining globally consistent spatial awareness in complex indoor layouts remains difficult, limiting their effectiveness on challenging VLN tasks.

Recently, several works~\cite{zeng2025janusvln,zheng2025efficient} have attempted to incorporate explicit 3D cues into MLLM-based navigation systems. However, such approaches typically reuse off-the-shelf geometry foundation models and do not fundamentally optimize geometric representations for the spatio-temporal dynamics inherent in navigation. Moreover, static geometric representations produced by existing foundation models~\cite{wang2025vggt,wang2025pi3} can degrade substantially under pronounced scene dynamics (e.g., highly dynamic human activities), which is particularly problematic for real-world VLN. In contrast to previous works that treat geometry as a plug-and-play module, our approach develops a dynamic geometry foundation model and infuses it into the VLN framework. This framework achieves more robust spatial reasoning and strong navigation performance in dynamic real-world environments.

%% file: vln_method.tex
We focus on vision-language navigation in continuous environments~\cite{krantz2020beyond} with monocular RGB observations. At each timestep $t$, the embodied agent is given an instruction $\mathcal{W}$ and an ego-centric RGB stream $\mathcal{O}_t = \{x_0, \dots, x_t\}$, where each frame $x_\tau$ denotes the observation at timestamp $\tau$. The agent’s goal is to learn a policy $\pi(\mathcal{O}_t, \mathcal{W})$ that maps the instruction and the history of visual observations to a low-level action $a_{t+1} \in \mathcal{A}$ for the next step. The agent operates in an atomic-level action space $\mathcal{A}$ composed of four actions: \{\texttt{Move Forward}, \texttt{Turn Left}, \texttt{Turn Right}, \texttt{Stop}\}. After executing $a_{t+1}$, the agent observes a new RGB frame $x_{t+1}$ and repeats this perception–action cycle until it gets the \texttt{Stop}.

The overall framework of our DyGeoVLN is illustrated in Fig~\ref{fig:vlnsys}. We first introduce our dynamic geometry foundation model (Section~\ref{subsec:dyamvgfm}), followed by the cross-branch feature fusion VLN architecture (Section~\ref{subsec:dualfusion}), and the adaptive-resolution and occupancy-aware spatial token pruning strategy (Section~\ref{subsec:tokenprune}).

\subsection{Dynamic Geometry Foundation Model}
\label{subsec:dyamvgfm}

\begin{figure}[!t]
\centering
\includegraphics[width=\linewidth]{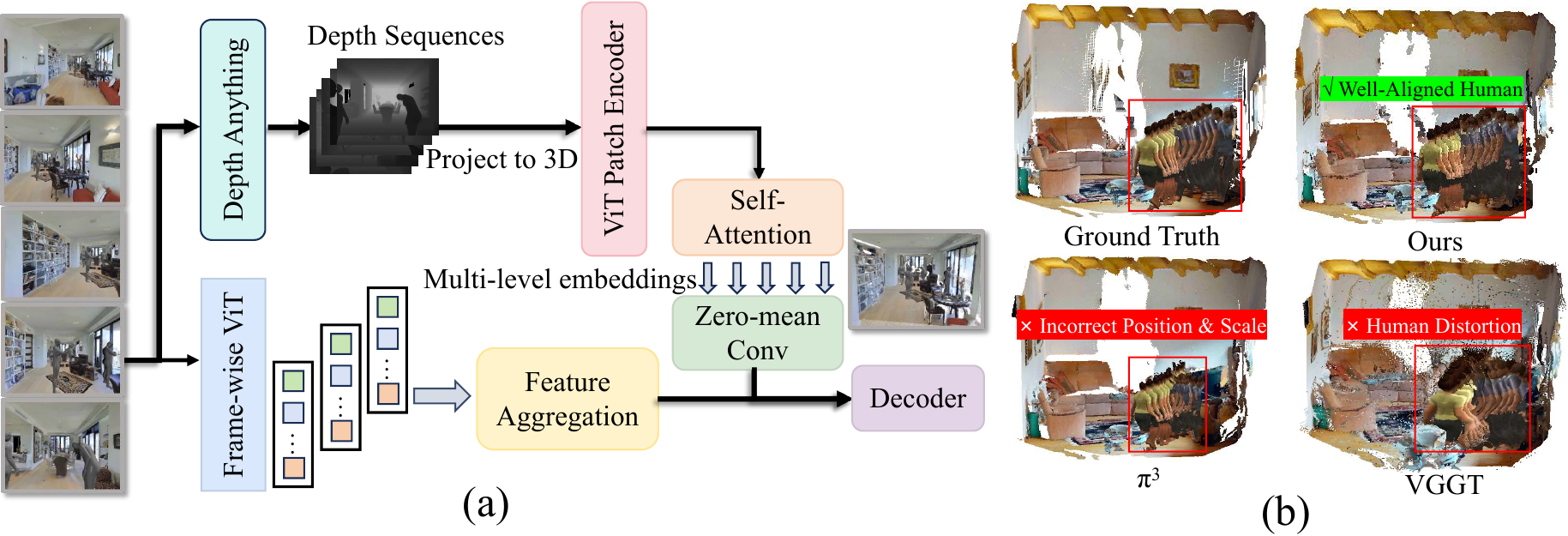}
\caption{Overview of the proposed DGFM. (a) The architecture of our dynamic geometry foundation model: 3D latent representations and 2D visual semantic features are fused and subsequently aligned along the temporal axis to achieve globally consistent and fine-grained reconstruction. (b) Qualitative reconstruction results compared against Ground Truth, $\pi^3$~\cite{wang2025pi3}, and VGGT~\cite{wang2025vggt}. The red bounding boxes highlight regions where both $\pi^3$ and VGGT suffer from poor human reconstruction. In contrast, our method preserves precise human positioning and maintains stable geometric structures.}
\label{fig:dynamicrecon}
\end{figure}

Recent visual geometry foundation models~\cite{wang2025vggt,wang2025pi3} primarily learn a direct mapping from 2D images to explicit 3D scene representations using large-scale multi-view supervision. Although such models achieve strong performance, they implicitly learn appearance and structure in the latent space and have to infer cross-frame consistency solely from RGB cues. This makes it difficult for the model to align the 3D structures of observations captured at different time steps, often leading to temporally inconsistent reconstructions and imperfect localization of dynamic objects. To address these issues, we propose a \textbf{dynamic geometry foundation model} that explicitly injects fine-grained depth-guided 3D information into the backbone, as illustrated in Fig.~\ref{fig:dynamicrecon}. Compared with~\cite{wang2025vggt,wang2025pi3}, our main technical novelty lies in multi-level 3D embeddings, dynamic-aware fusion with visual geometry backbone and hierarchical latent decoding, which are tailored to dynamic environments.

\noindent\textbf{Depth-guided local point maps.}
Given a sequence of monocular images $\{I_t\}_{t=1}^{T}$, we first predict a dense depth map $D_t$ for each frame using an off-the-shelf depth estimator (Depth Anything~\cite{yang2024depth}). For every valid pixel location $\mathbf{u} = (u,v)$ in $I_t$, we project it into 3D coordinates using the intrinsic matrix $\mathbf{K}$: $\mathbf{p}_t(\mathbf{u}) = D_t(\mathbf{u})\, \mathbf{K}^{-1} \tilde{\mathbf{u}},$
where $\tilde{\mathbf{u}} = (u,v,1)^\top$ is the homogeneous pixel coordinate. The $\mathbf{P}_t = \{\mathbf{p}_t(\mathbf{u})\}$ forms a local point map anchored at the current camera.

\noindent\textbf{Point-map Transformer and multi-level 3D embeddings.}
The point map $\mathbf{P}_t$ is discretized into a regular grid and fed to a point-map Transformer encoder. Specifically, we first partition $\mathbf{P}_t$ into non-overlapping patches and embed each patch into a token $\mathbf{Z}_t^{(0)} = \phi_{\text{patch}}(\mathbf{P}_t)$. A stack of self-attention blocks with coarse-to-fine increasing receptive fields is then applied to obtain a hierarchical 3D representations
\begin{equation}
    \mathbf{G}_t^{(1)}, \mathbf{G}_t^{(2)}, \ldots, \mathbf{G}_t^{(S)} 
    = f_{\text{transformer}}(\mathbf{Z}_t^{(0)}),
\end{equation}
where $\mathbf{G}_t^{(s)}$ captures 3D context at the $s$-th granularity level. These multi-level 3D embeddings serve as explicit geometric representations that complement the structural features of the visual backbone.

\noindent\textbf{Dynamic-aware fusion with visual geometry backbone.}
In parallel to the 3D branch, we reuse a frame-wise ViT encoder~\cite{wang2025vggt} to extract 2D visual tokens~\cite{oquab2023dinov2} from each RGB frame, followed by the original feature aggregation module of the geometric foundation model~\cite{wang2025pi3} to obtain per-layer visual feature maps $\{\mathbf{E}_t^{(s)}\}_{s=1}^{S}$. To inject our 3D embeddings into the backbone without destroying its pre-trained parameter distribution, we adopt a zero-mean convolutional module. For each layer $s$, we first align the channel shape of $\mathbf{G}_t^{(s)}$ and then fuse it with $\mathbf{E}_t^{(s)}$ through
\begin{equation}
    \tilde{\mathbf{E}}_t^{(s)} = g_{\text{zm}}\bigl(\mathbf{G}_t^{(s)}\bigr) + \mathbf{E}_t^{(s)}, \quad s = 1,\ldots,S,
\end{equation}
where $g_{\text{zm}}(\cdot)$ is a convolution layer whose weights and bias are initialized to zero. During fine-tuning, the module learns to gradually inject depth-guided geometric cues as a residual term, enabling the model to benefit from explicit 3D representation information while preserving the internal prior weights encoded in the original foundation model.

\noindent\textbf{Decoder and dynamic 3D reconstruction.} 
The fused features $\{\tilde{\mathbf{E}}_t^{(s)}\}_{s=1}^{S}$ are fed into the pre-trained geometric decoder, which produces dense 3D scene predictions such as camera poses, global point clouds, or local point maps, depending on the underlying attention head types. We first concatenate the multi-level fused features of each frame as $\tilde{\mathbf{E}}_t = [\tilde{\mathbf{E}}_t^{(1)}, \ldots, \tilde{\mathbf{E}}_t^{(S)}]$ and pass the sequence $\{\tilde{\mathbf{E}}_t\}_{t=1}^{T}$ through the decoder $\mathcal{D}$ to get the decoded geometric token map $\{\mathbf{H}_t\}_{t=1}^{T} = \mathcal{D}\bigl(\{\tilde{\mathbf{E}}_t\}_{t=1}^{T}\bigr),$
\noindent To explicitly structure the latent geometry at different scales, we further perform a hierarchical latent decoding on top of the shared tokens $\mathbf{H}_t$. We first employ two dedicated decoders to obtain geometry embeddings for camera and local points: $    \{\mathbf{H}_t^{\text{cam}}\}_{t=1}^{T}
    = \mathcal{D}_{\text{cam}}\bigl(\{\mathbf{H}_t\}_{t=1}^{T}\bigr),
    \{\mathbf{H}_t^{\text{loc}}\}_{t=1}^{T}
    = \mathcal{D}_{\text{loc}}\bigl(\{\mathbf{H}_t\}_{t=1}^{T}\bigr),$
where $\mathbf{H}_t^{\text{cam}}$ and $\mathbf{H}_t^{\text{loc}}$ are latent geometry representations specialized for camera motion and local point-wise structure. We then aggregate these latent embeddings as the conditioning signal for the global point decoder. Denoting a learnable aggregation operator by $\mathcal{A}(\cdot)$, we form $\mathbf{U}_t =\mathcal{A}\Bigl(\mathbf{H}_t^{\text{cam}}, \mathbf{H}_t^{\text{loc}}\Bigr)$ and decode global-scale latent representations via
\begin{equation}
    \{\mathbf{H}_t^{\text{glo}}\}_{t=1}^{T}
    = \mathcal{D}_{\text{glo}}\bigl(\{\mathbf{U}_t\}_{t=1}^{T}\bigr),
    \label{eq:latent-global}
\end{equation}
where $\mathbf{H}_t^{\text{glo}}$ captures global-scale latent geometry that is explicitly conditioned on the latent camera and local point information.
\noindent Finally, task-specific prediction heads project the corresponding hidden states to explicit 3D quantities.
Because the shared decoder receives embeddings that jointly encode visual appearance, local point-wise geometry, and multi-scale 3D context, and the global branch is further conditioned on the latent camera and local representations, the model is encouraged to produce geometry tokens and dense reconstructions that are metrically consistent across time and aligned between different timestamps. This dynamic geometry foundation model serves as the 3D encoder in DyGeoVLN, providing reliable geometric tokens for navigation tasks.

\begin{figure}[!t]
\centering
\includegraphics[width=\linewidth]{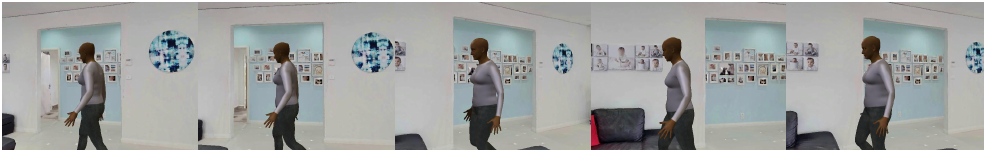}
\caption{Example case of our DyHM3D dataset.}
\label{fig:DynHM3D dataset}
\end{figure}

\noindent\textbf{Construction of dynamic 3D geometry dataset.}
Training geometry foundation models for dynamic perception in support of VLN requires 3D datasets with dynamic scenes. However, existing 3D geometry datasets used for VLN mainly focus on static environments. To address this limitation, we propose DyHM3D, a dynamic human-centric 3D dataset built on the Habitat-Matterport 3D Dataset (HM3D). Since humans are the most common dynamic entities in indoor VLN scenarios, we focus on simulating human movements. In theory, our dataset construction pipeline can also be extended to other types of dynamic objects. Specifically, we first sample human motion waypoints within navigable regions on the navigation mesh provided by Habitat Simulator. We then employ skeletal-driven 3D human models with motion controllers to generate natural walking motions along these paths. Meanwhile, we adopt a camera-following strategy based on linear interpolation between adjacent waypoints, constraining the camera to the navigation mesh to ensure smooth and physically consistent motion. Data are collected from a first-person perspective under diverse observation settings, including multiple camera heights and horizontal fields of view. For each trajectory, RGB images, depth maps, camera poses, and camera intrinsics are recorded synchronously. In total, DyHM3D contains approximately 50,000 trajectories, as shown in Fig.~\ref{fig:DynHM3D dataset}.

\subsection{Cross-branch Feature Fusion VLN architecture}
\label{subsec:dualfusion}

To enable the agent with both semantic understanding and spatial reasoning, we design a cross-branch fusion architecture that processes each observation frame with a 2D semantic branch and a 3D spatial-geometric branch, and then fuses them into unified visual–spatial tokens, as illustrated in Fig.~\ref{fig:vlnsys}. The 2D branch inherits from the vision encoder in Qwen2-VL~\cite{wang2024qwen2}, while the 3D branch is instantiated by our dynamic geometry foundation model introduced in Sec.~\ref{subsec:dyamvgfm}.

\noindent\textbf{2D semantic tokenization.}
Given an RGB frame $x_t$, the Qwen-VL visual encoder divides the image into non-overlapping $p_v \times p_v$ patches and maps each patch to a visual token $\mathbf{V}_t = f_{\text{encode}}(x_t), \;
\mathbf{V}_t \in \mathbb{R}^{L_v \times C_v}, \; L_v = \tfrac{H}{p_v} \times \tfrac{W}{p_v}$. To interface with the language backbone, we use a multi-modal projector $f_{\text{multi-modal}}$ to map the visual tokens $\mathbf{V}_t$ into the embedding space $\mathbf{X}_t$.

\noindent\textbf{3D spatial-geometric tokenization.}
In parallel, the dynamic geometry foundation model takes the same frame $x_t$ and its depth-guided local point map as input, and outputs multi-level 3D embeddings (Sec.~\ref{subsec:dyamvgfm}). We aggregate these embeddings into a sequence of 3D tokens $\mathbf{H}_t^{\text{glo}} = f_{\text{3D}}(x_t), \; 
\mathbf{H}_t^{\text{glo}} \in \mathbb{R}^{L_g \times C_g}$. We then use a feature alignment $f_{\text{align}}$ to project $\mathbf{H}_t^{\text{glo}}$ into $\tilde{\mathbf{H}}_t^{\text{glo}}$, which has the same dimensionality as the visual tokens.

\noindent\textbf{Cross-branch 2D–3D fusion.}
After obtaining visual semantic tokens $\mathbf{X}_t$ and spatial tokens $\mathbf{G}_t$, we fuse them using a multi-head cross-attention module:
\begin{equation}
    \mathbf{F}_t = \text{Cross-Atten}\bigl(Q=\mathbf{X}_t,\; K=\tilde{\mathbf{H}}_t^{\text{glo}},\; V=\tilde{\mathbf{H}}_t^{\text{glo}}),
\end{equation}
where $\text{Cross-Atten}(\cdot)$ denotes a multi-head cross-attention layer and 
$\mathbf{F}_t$ is the fused visual–spatial token sequence fed into the LLM module. In this way, each 2D patch token in $\mathbf{X}_t$ incorporates global spatial latent representation in $\tilde{\mathbf{H}}_t^{\text{glo}}$, producing fused tokens $\mathbf{F}_t$ that preserve the spatial structure.

\noindent\textbf{Sliding-window KV cache for active frames.}
During navigation, we use an active sliding window to select the most recent $N$ frames. For each step $i$, we define the active frame index set $\Omega_i$ and compute fused tokens $\{\mathbf{F}_t\}_{t \in \Omega_i}$ for all frames in the window. These tokens, together with instruction tokens and pruned memory tokens from inactive windows, are fed into the decoder to predict the following timestamp action $a_i$:
\begin{equation}
    a_i = \text{Decoder}\bigl(\mathcal{I}, \{\mathbf{M}_k\}, \{\mathbf{F}_t\}_{t \in \Omega_i}; \mathcal{KV}_i^{\text{window}}\bigr),
\end{equation}
where $\mathcal{I}$ denotes the instruction tokens, $\mathbf{M}_k$ are spatially pruned memory tokens from historical frames, and $\mathcal{KV}_i^{\text{window}}$ denotes the Key/Value cache associated with the current sliding window. At each step we append the KV states of the newest frame and remove the KV states of the oldest frame once it falls outside the $N$ frame window. This sliding-window KV cache reuses attention projections for recent observations while keeping the effective context length, thereby making memory and runtime bounded throughout navigation process.

\subsection{Adaptive-resolution and occupancy-aware spatial token pruning}
\label{subsec:tokenprune}
Long-horizon navigation inevitably accumulates a large number of tokens. If all tokens are fed into the LLM, the context window is quickly saturated, causing inference to become progressively slower. To address this problem, we design an adaptive-resolution and occupancy-aware spatial token pruning module that 
eliminates spatio-temporal redundancy. To preserve geometrically informative tokens and enhance the token temporal consistency, we propose an importance-aware token completion strategy and a temporal smoothing method.
Different from previous voxel-based pruning methods~\cite{wei2025streamvln} that rely on perfect depth images and externally estimated camera poses captured from Habitat~\cite{savva2019habitat}, 
our approach is pose-free and depth-free from the sensor perspective: the dynamic geometry foundation model directly outputs a local point map and the corresponding camera-to-world pose for each frame. For detailed algorithm working flow, please see our supplementary material.

\noindent\textbf{Adaptive-resolution voxel grouping.} We design an adaptive-resolution strategy when assigning tokens to voxels. During projection, we estimate the depth information of valid points in each frame and adjust the base voxel size accordingly. In addition, each token is further modulated by a depth-dependent scale factor: tokens with small depth are grouped with a smaller scale, mid-range tokens with a moderate scale, and far-range tokens with a larger scale. This two-level adaptation allows us to retain more tokens around nearby geometry while aggressively merging and pruning redundant tokens in distant regions.

\noindent\textbf{Occupancy-aware spatial token pruning.} After voxelization, multiple tokens from different frames may occupy the same voxel cell. For each voxel, our module supports several selection rules: (i) a \textit{latest} rule that keeps the token from the most recent frame; (ii) a \textit{priority} rule that ranks tokens by a weighted combination of recency and proximity (closer and newer observations are preferred); and 
(iii) a \textit{multi-token} rule that preserves the top-$K$ tokens in each voxel to balance redundancy reduction and diversity. In all cases, only the selected tokens are kept as occupancy representatives of that voxel; all other tokens mapped to the same cell are masked out. This occupancy-aware pruning ensures that each voxel contributes an informative set of tokens describing the visible surfaces within it.

\noindent\textbf{Importance-aware token completion.}
In sparse regions, voxel pruning yields too few tokens, thus harming the model's robustness. To address this, we monitor the number of preserved tokens per frame and enforce a minimum keep ratio $\rho$. If the number of selected tokens is lower than $\rho$, we compute an importance score for each discarded token and re-calculate the top-scoring ones until the ratio is satisfied. The importance score is a weighted sum of four factors: feature magnitude, range information, spatial distribution, and temporal recency. This strategy ensures that even aggressively pruned frames can still retain a sufficient number of semantically and geometrically salient tokens.

\noindent\textbf{Temporal smoothing of pruning masks.} To enhance temporal consistency, we apply temporal smoothing to the binary pruning masks. For each frame, we collect masks from a short temporal window (e.g., the current frame and its two neighbors) and perform a majority vote at each token index. Tokens that are consistently preserved across the window remain active, while isolated fluctuations are marked as inactive.

%% file: gdpf_exp.tex
\subsection{Experimental Setup}
\noindent\textbf{Simulation Benchmark and Evaluation Metrics.} We simultaneously evaluate our method on both dynamic and static benchmarks. For the dynamic benchmark, we use the most dominant dynamic benchmark HA-VLN~\cite{dong2025ha}, which contains multi-human interactions, fine-grained language-motion alignment, and explicit social modeling. The evaluation metrics used for the dynamic HA-VLN benchmark are Total Collision Rate (TCR): quantifies how often collisions occur in human-occupied zones; Collision Rate per step (CR): the fraction of navigation instances incurring at least one collision; Navigation Error (NE): the mean distance between agent’s final position and intended target; Success Rate (SR): measures the ratio of successful completed episodes. For the static benchmark, we evaluate our model on R2R-CE~\cite{krantz2020beyond} benchmark using the Habitat simulator~\cite{savva2019habitat}. For R2R-CE evaluation metrics, we adopt widely-used metrics~\cite{mattersim,anderson2018evaluation,ilharco2019general}, consisting of navigation error (NE): average geometric distance in meters between the final and ground-truth location; success rate (SR): the ratio of navigation paths with NE less than 3 meters; oracle SR (OSR): SR based on the stop policy; SR penalized by path length (SPL). 


\noindent\textbf{Real-world Setup and Evaluation Metric.} Our model is evaluated in various real-world settings, including corridors, lobbies, indoor rooms, and crowded environments. We compare our DyGeoVLN with NaVid~\cite{zhang2024navid}, Uni-NaVid~\cite{zhang2024uni}, NaVILA~\cite{cheng2025navila}, and StreamVLN~\cite{wei2025streamvln}. For each scenario, we evaluated 20 episodes per model, using Success Rate (SR) as the primary evaluation metric. For the real-world experiments, we build our DyGeoVLN model with diffusion action head following~\cite{wei2025groundslowfastdualsystem}. The high-level system is our DyGeoVLN model, which performs geometry and dynamic-aware spatial reasoning over streaming RGB observations and instructions, and outputs an image-pixel goal indicating the target location in the current image view. The low-level system is a diffusion action module that takes the predicted pixel goal and RGB images as input and generates continuous trajectories toward the goal. The resulting trajectories are converted into control inputs by an MPC controller and sent to the Unitree Go1. 



\noindent\textbf{Implementation Details.} 
We train our DGFM on our proposed DyHM3D dataset. The model weights are initialized using a two-part strategy: the frame-wise ViT encoder, camera decoder, and local point decoder inherit their initial weights from $\pi^3$~\cite{wang2025pi3}, whereas our multi-level 3D embeddings, dynamic-aware fusion, and global-scale latent representations are trained from scratch. During the training of DyGeoVLN, the vision encoder and the large language model are initialized with pre-trained weights from~\cite{wang2024qwen2} and subsequently updated, whereas the DGFM remains frozen. For training dataset, we use R2R-CE~\cite{krantz2020beyond}, RxR-CE~\cite{ku2020room}, EnvDrop~\cite{tan2019learning}, HA-VLN~\cite{dong2025ha}, and ScaleVLN~\cite{wang2023scalevln}. We then use the DAgger~\cite{ross2011reduction} algorithm to enhance the model's generalization and adaptation to novel scenes and imperfect trajectories.

\subsection{Experiment Result}

\begin{figure*}[!t]
\centering
    \includegraphics[width=\linewidth]{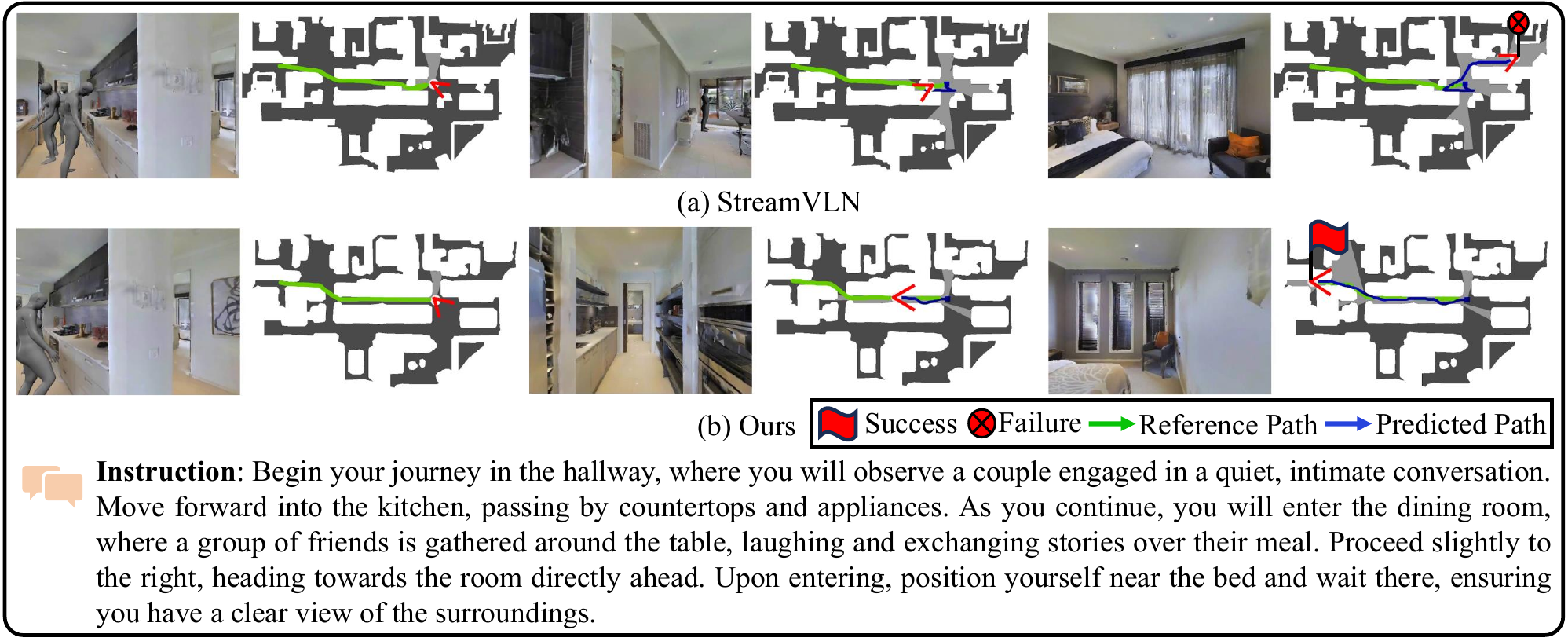}
    \caption{Qualitative comparison between state-of-the-art method StreamVLN~\cite{wei2025streamvln} and our DyGeoVLN on the dynamic HA-VLN~\cite{dong2025ha} benchmark. StreamVLN loses track of the path and fails to handle dynamic human context, while DyGeoVLN better understands human behaviors and accurately reaches the target destination.}
    \label{fig:har2r_compare}
\end{figure*}

\noindent\textbf{Results on Dynamic VLN Benchmark.} Table~\ref{table:har2r-ce} reports the navigation performance of our method on the dynamic HA-VLN benchmark. We also provide a qualitative comparison in Fig.~\ref{fig:har2r_compare}. Compared with the stationary VLN benchmark, this benchmark is more dynamic and crowded. Compared with g3D-LF~\cite{wang2025g3d}, which is trained with panorama RGB-D observations and large-scale additional 3D structural data, our method achieves consistent gains on both splits: on ``Val-Seen'', we obtain a $12\%$ improvement in SR and a $10\%$ reduction in CR; on ``Val-Unseen'', we achieve an $13\%$ improvement in SR and a $11\%$ reduction in CR. Compared with VLM-based methods~\cite{wei2025streamvln,cheng2025navila}, we achieve an average $10\%$ gain in SR and a $8\%$ reduction in CR. This improvement comes from the strong dynamic scene representation of our geometry foundation model. By integrating geometric and semantic tokens, the model achieves stronger spatial understanding and more reliable navigation in dynamic environments. We provide additional examples in the supplementary material to further demonstrate the effectiveness of our dynamic geometry foundation model.


\input{newtable}

\begin{figure*}[!t]
\centering
    \includegraphics[width=\linewidth]{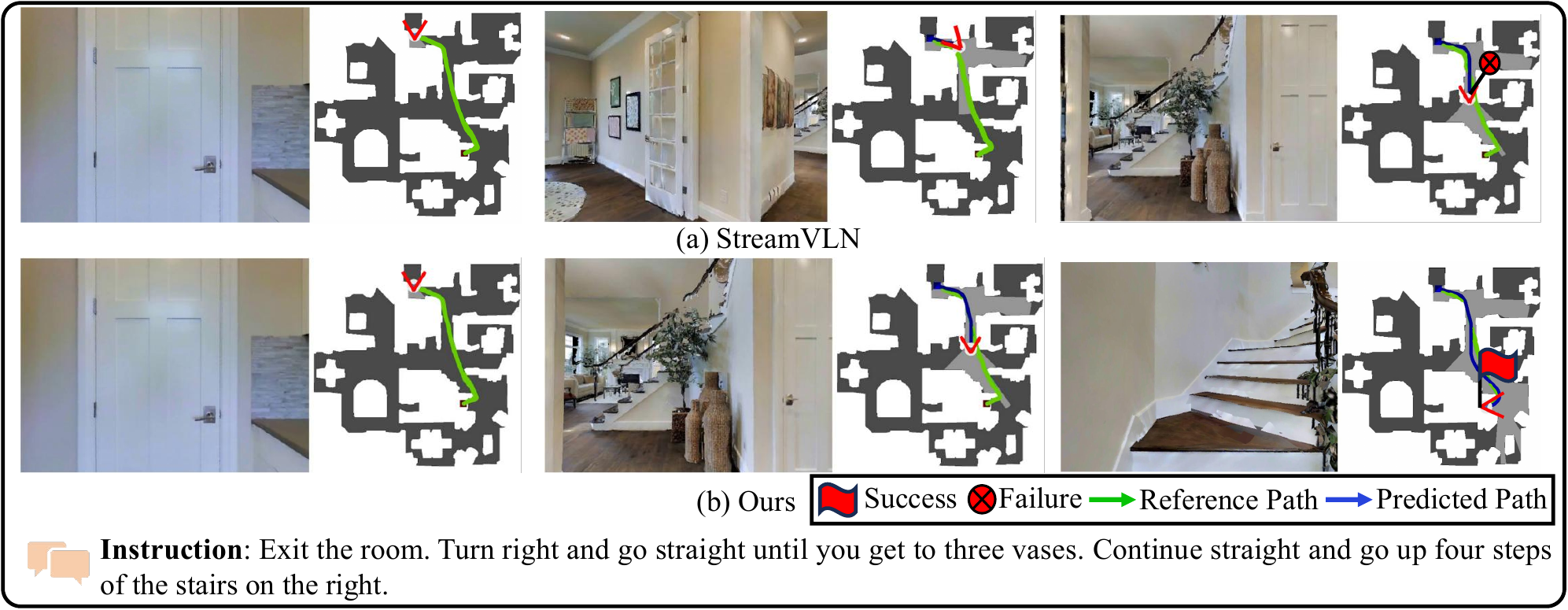}
    \caption{Qualitative comparison between state-of-the-art method StreamVLN~\cite{wei2025streamvln} and our DyGeoVLN on the R2R-CE~\cite{krantz2020beyond} benchmark. DyGeoVLN demonstrates improved trajectory consistency, whereas StreamVLN stops midway and fails to reach the destination.}
    \label{fig:r2r_compare}
\end{figure*}

\noindent\textbf{Results on Static VLN Benchmark.} Table~\ref{table:vlm-r2r-ce} shows the navigation performance of our method on the R2R-CE benchmark in terms of ``Val-Unseen'' split. We also provide a qualitative comparison in Fig.~\ref{fig:r2r_compare}. Our DyGeoVLN model achieves state-of-the-art performance compared with both Monocular RGB-only methods and Panoramic RGB-D/Odometry-based methods. Our method achieves $4.41$ NE, $70.1\%$ OSR, $60.8\%$ SR, and $55.8\%$ SPL, setting a new best performance among monocular RGB-only methods. Moreover, despite relying solely on monocular RGB observations, our method surpasses state-of-the-art panoramic RGB-D/odometry-based models, such as g3D-LF~\cite{wang2025g3d} and ETPNav~\cite{an2024etpnav}, across all four metrics, further demonstrating strong generalization and robust performance on static VLN benchmark.

\begin{table*}[!b]
\renewcommand\arraystretch{1.0}
\setlength{\tabcolsep}{7pt} 
\centering
\caption{Comparison with state-of-the-art methods on R2R-CE~\cite{krantz2020beyond} dataset. $*$ indicates methods using the waypoint predictor from ~\cite{hong2022bridging}. Complex Data indicates the method simultaneously uses panorama RGB-D images and/or odometry. Simple Data indicates the method only uses Monocular RGB images.}
\label{table:vlm-r2r-ce}

\scriptsize 

\begin{tabular}{l|cc|cccc}
\toprule
\multirow{2}{*}{Methods} &
\multicolumn{2}{c|}{\textsf{Observation Data}} &
\multicolumn{4}{c}{\textsf{R2R-CE Val-Unseen}} \\
\cmidrule(lr){2-3}\cmidrule(lr){4-7}
 & Complex Data & Simple Data
 & NE$\downarrow$ & OSR$\uparrow$ & SR$\uparrow$ & SPL$\uparrow$ \\
\midrule
GridMM$^{*}$~\cite{wang2023gridmm}
 & \cmark & 
 & 5.11 & 61.0 & 49.0 & 41.0 \\
ETPNav$^{*}$~\cite{an2024etpnav}
 & \cmark & 
 & 4.71 & 65.0 & 57.0 & 49.0 \\
ScaleVLN$^{*}$~\cite{wang2023scalevln}
 & \cmark & 
 & 4.80 & 64.0 & 55.0 & 51.0 \\
g3D-LF$^{*}$~\cite{wang2025g3d}
 & \cmark & 
 & 4.53 & 68.0 & 61.0 & 52.0 \\
CM2~\cite{georgakis2022cross}
 & \cmark & 
 & 7.02 & 41.5 & 34.3 & 27.6 \\
WS-MGMap~\cite{chen2022weakly}
 & \cmark & 
 & 6.28 & 47.6 & 38.9 & 34.3 \\
Sim2Real-3DFF$^{*}$~\cite{2024-arxiv-Sim-To-Real}
 & \cmark & 
 & 5.95 & 55.8 & 44.9 & 30.4 \\
NavMorph~\cite{yao2025navmorph}
 & \cmark & 
 & 5.75 & 56.9 & 47.9 & 33.2 \\
MapNav~\cite{zhang2025mapnav}
 & \cmark & 
 & 4.93 & 53.0 & 39.7 & 37.2 \\
Dynam3D~\cite{wang2025dynam3d}
 & \cmark & 
 & 5.34 & 62.1 & 52.9 & 45.7 \\
NaVid~\cite{zhang2024navid}
 &  & \cmark
 & 5.47 & 49.1 & 37.4 & 35.9 \\
Uni-NaVid~\cite{zhang2024uni}
 &  & \cmark
 & 5.58 & 53.3 & 47.0 & 42.7 \\
NaVILA~\cite{cheng2025navila}
 &  & \cmark
 & 5.22 & 62.5 & 54.0 & 49.0 \\
StreamVLN~\cite{wei2025streamvln}
 &  & \cmark
 & 5.10 & 64.0 & 55.7 & 50.9 \\
NavFoM~\cite{zhang2025embodied}
 &  & \cmark
 & 5.01 & 64.9 & 56.2 & 51.2 \\
\textbf{DyGeoVLN (ours)}
 &  & \cmark
 & \textbf{4.41} & \textbf{70.1} & \textbf{60.8} & \textbf{55.8} \\
\bottomrule
\end{tabular}
\end{table*}

\begin{figure}[!t]
\centering
\includegraphics[width=\linewidth]{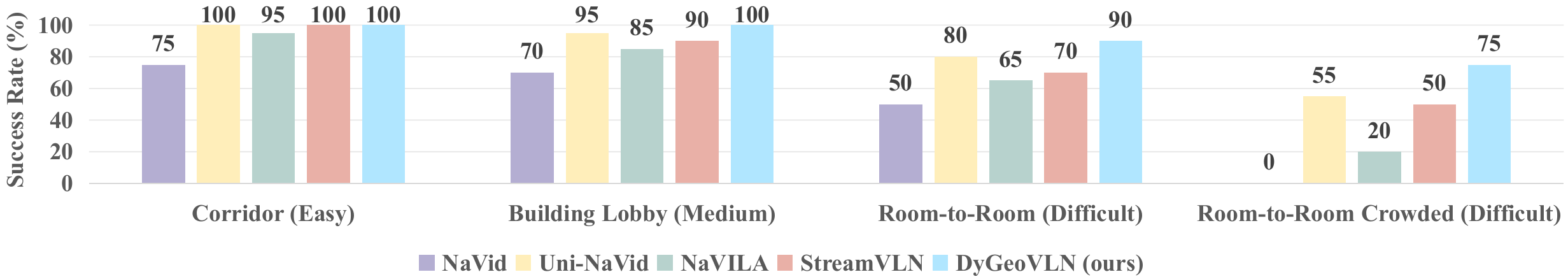}
\caption{Quantitative comparison of success rates (\%) across varying real-world scenarios. Our proposed DyGeoVLN consistently outperforms other baseline models, demonstrating particular robustness across diverse environments.}
\label{fig:real}
\end{figure}

\begin{figure}[!t]
\centering
\includegraphics[width=\linewidth]{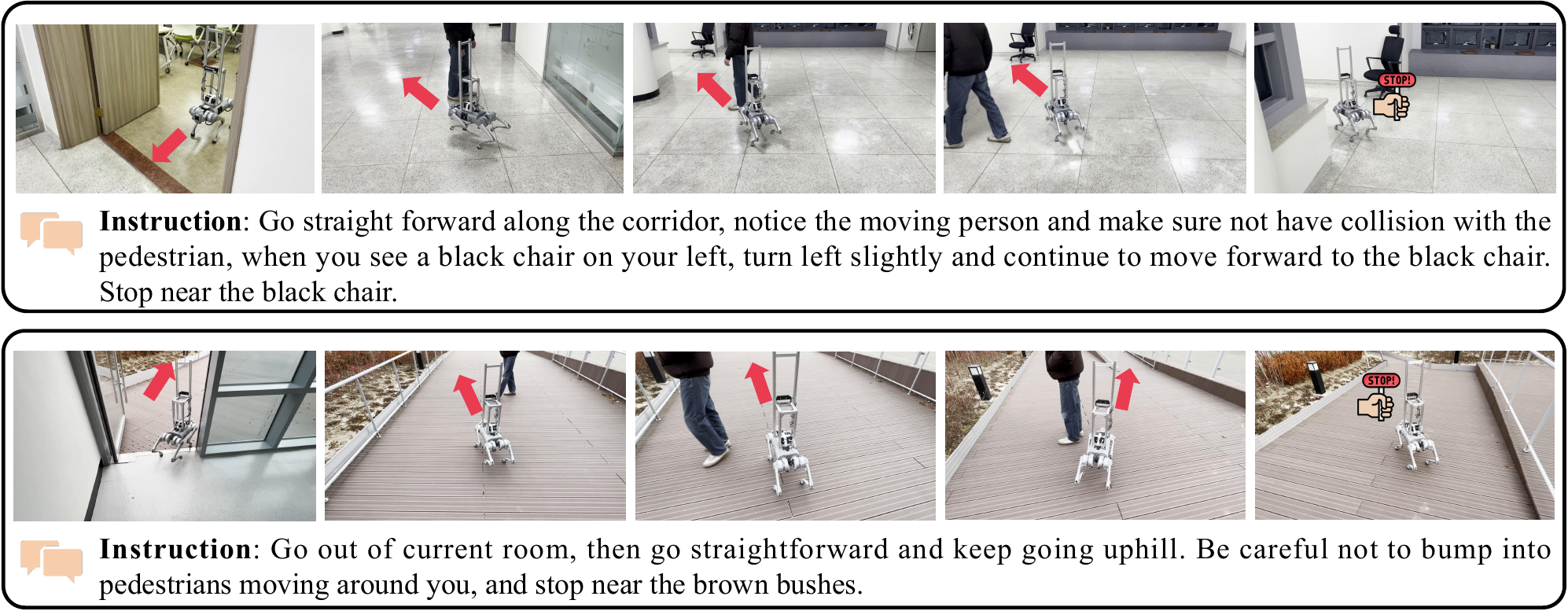}
\caption{Qualitative results of our DyGeoVLN model. The robot successfully follows instructions to reach the goal while avoiding collisions with moving pedestrians.}
\label{fig:realexp}
\end{figure}

\noindent\textbf{Real-world Results.} As shown in the quantitative results in Fig.~\ref{fig:real}, our DyGeoVLN achieves robust performance at all difficulty levels. Fig.~\ref{fig:realexp} provides qualitative demonstrations, showing that our model can perform tasks effectively in dynamic environments with moving pedestrians. Additional results are detailed in the supplementary material.

\subsection{Ablation Study}

We conducted the ablation study on the dynamic HA-VLN~\cite{dong2025ha} benchmark to validate the effect of our proposed DyGeoVLN. We report navigation performance on the ``Val-Unseen'' split of the HA-VLN benchmark, as shown in Table~\ref{table:ablation_final}.

\noindent\textbf{Cross-Branch Fusion Strategy.} We first investigate the ablation study of our dual-branch architecture design. Using only the 2D vision encoder or the 3D geometry encoder resulted in a significant decrease in the model's navigation performance, suggesting that only coupling 2D visual semantic features and 3D spatial geometric features together can produce the best result.

\noindent\textbf{Explicit Dynamic-aware 3D Spatial Injection.} We further examine the effect of our explicit 3D spatial information injection in our dynamic geometry foundation model. The success rate of our method improves $4\%$, and the collision rate reduces $5\%$ with the help of an explicit injection of 3D information. This shows that our dynamic geometry foundation model can improve navigation performance.

\noindent\textbf{Token Pruning.} We also evaluate the impact of our adaptive-resolution and occupancy-aware spatial token pruning strategy. With token reduction, the success rate of the model increases by $3\%$, while the collision rate and navigation error decrease, validating that the removal of redundant temporal and spatial tokens helps our model focus more on the critical region and improves navigation performance.

\begin{table}[!t]
    \centering
    \scriptsize
    \setlength{\tabcolsep}{6pt}
    \renewcommand\arraystretch{1}
    \caption{Ablation Study of each component in our proposed DyGeoVLN.}
    \label{table:ablation_final}
    \setlength{\tabcolsep}{10pt} 
    \begin{tabular}{l|cccc}
        \hline
        Methods & NE$\downarrow$ & TCR$\downarrow$ & CR$\downarrow$ & SR$\uparrow$ \\
        \hline
        \textbf{DyGeoVLN (Full)} & \textbf{5.12} & \textbf{3.69} & \textbf{0.38} & \textbf{0.40} \\
        \hline
        w/o Visual Semantic & 6.82 & 4.96 & 0.51 & 0.30 \\
        w/o Spatial Geometry & 5.54 & 4.03 & 0.43 & 0.34 \\
        w/o Dynamic Spatial Injection & 5.37 & 4.10 & 0.43 & 0.36 \\
        w/o Spatial Token Pruning & 5.33 & 3.94 & 0.39 & 0.37 \\
        \hline
    \end{tabular}
\end{table}

%% file: newtable.tex
\begin{table}[!b]
\renewcommand\arraystretch{1.0}
\scriptsize 
\centering
\caption{Comparison with state-of-the-art methods on dynamic HA-VLN~\cite{dong2025ha} benchmark. $*$ indicates methods using the pretrained waypoint predictor from~\cite{hong2022bridging}. Complex Data indicates the method simultaneously uses panorama RGB-D images and/or odometry. Simple Data indicates the method only uses Monocular RGB images.}
\resizebox{\textwidth}{!}{ 
\begin{tabular}{l|cc|cccc|cccc}
\toprule
\multirow{2}{*}{Methods} &
\multicolumn{2}{c|}{\textsf{Observation Data}} &
\multicolumn{4}{c}{\textsf{Val Seen}} & \multicolumn{4}{c}{\textsf{Val Unseen}}\\
\cmidrule(lr){2-3}\cmidrule(lr){4-7}\cmidrule(lr){8-11}
 & Complex Data & Simple Data
 & NE$\downarrow$ & TCR$\downarrow$ & CR$\downarrow$ & SR$\uparrow$ & NE$\downarrow$ & TCR$\downarrow$ & CR$\downarrow$ & SR$\uparrow$ \\
\midrule
VLN-CMA~\cite{krantz2020beyond} & \cmark &  & 7.63 & 63.09 & 0.75 & 0.04 & 7.34 & 47.06 & 0.77 & 0.07 \\
HA-VLN-VL~\cite{dong2025ha} & \cmark &  & 5.02 & 4.44 & 0.52 & 0.20 & 5.25 & 6.63 & 0.59 & 0.14 \\
BEVBert$^{*}$~\cite{an2023bevbert} & \cmark &  & 5.53 & 3.64 & 0.46 & 0.27 & 5.51 & 4.71 & 0.55 & 0.21 \\
ETPNav$^{*}$~\cite{an2024etpnav} & \cmark &  & 5.17 & 4.07 & 0.43 & 0.24 & 5.43 & 6.94 & 0.58 & 0.17 \\
g3D-LF$^{*}$~\cite{wang2025g3d} & \cmark &  & 5.12 & 3.58 & 0.41 & 0.32 & 5.30 & 4.54 & 0.49 & 0.27 \\
NaVid~\cite{zhang2024navid} &  & \cmark & 6.62 & 6.26 & 0.51 & 0.36 & 7.49 & 6.17 & 0.49 & 0.34 \\
Uni-NaVid~\cite{zhang2024uni} &  & \cmark & 6.27 & 5.96 & 0.50 & 0.37 & 7.74 & 6.45 & 0.55 & 0.32 \\
NaVILA~\cite{cheng2025navila} &  & \cmark & 5.95 & 3.86 & 0.42 & 0.33 & 6.39 & 4.42 & 0.45 & 0.32 \\
StreamVLN~\cite{wei2025streamvln}
 &  & \cmark
 & 5.52 & 3.72 & 0.41 & 0.34 & 5.59 & 4.03 & 0.42 & 0.33 \\
\textbf{DyGeoVLN (ours)} &  & \cmark & \textbf{4.78} & \textbf{3.11} & \textbf{0.31} & \textbf{0.44} & \textbf{5.12} & \textbf{3.69} & \textbf{0.38} & \textbf{0.40} \\
\bottomrule
\end{tabular}
\label{table:har2r-ce}
} %
\end{table}